# Recognizing Speech in a Novel Accent: The Motor Theory of Speech Perception Reframed

Clément Moulin-Frier and Michael A. Arbib

CONTENTS

## ABSTRACT

The motor theory of speech perception holds that we perceive the speech of another in terms of a motor representation of that speech. However, when we have learned to recognize a foreign accent, it seems plausible that recognition of a word rarely involves reconstruction of the speech gestures of the speaker rather than the listener. To better assess the motor theory and this observation, we proceed in three stages. Part 1 places the motor theory of speech perception in a larger framework based on our earlier models of the adaptive formation of mirror neurons for grasping, and for viewing extensions of that mirror system as part of a larger system for neuro-linguistic processing, augmented by the present consideration of recognizing speech in a novel accent. Part 2 then offers a novel computational model of how a listener comes to understand the speech of someone speaking the listener's native language with a foreign accent. The core tenet of the model is that the listener uses hypotheses about the word the speaker is currently uttering to update probabilities linking the sound produced by the speaker to phonemes in the native language repertoire of the listener. This, on average, improves the recognition of later words. This model is neutral regarding the nature of the representations it uses (motor vs. auditory). It serve as a reference point for the discussion in Part 3, which proposes a dual-stream neuro-linguistic architecture to revisits claims for and against the motor theory of speech perception and the relevance of mirror neurons, and extracts some implications for the reframing of the motor theory.

**TEXT**

# Recognizing Speech in a Novel Accent: The Motor Theory of Speech Perception Reframed

Clément Moulin-Frier

*GIPSA-Lab, Grenoble University*

*Inria/Ensta-Paristech, Flowers team, Bordeaux, France*

+33 6 62 56 42 89

Clement.moulin-frier@gmail.com

Michael A. Arbib

*USC Brain Project, University of Southern California*

(213) 740-9220

arbib@usc.edu

The motor theory of speech perception holds that we perceive the speech of another in terms of a motor representation of that speech. However, when we have learned to recognize a foreign accent, it seems plausible that recognition of a word rarely involves reconstruction of the speech gestures of the speaker rather than the listener. To better assess the motor theory and this observation, we proceed in three stages. Part 1 places the motor theory of speech perception in a larger framework based on our earlier models of the adaptive formation of mirror neurons for grasping, and for viewing extensions of that mirror system as part of a larger system for neuro-linguistic processing, augmented by the present consideration of recognizing speech in a novel accent. Part 2 then offers a novel computational model of how a listener comes to understand the speech of someone speaking the listener's native language with a foreign accent. The core tenet of the model is that the listener uses hypotheses about the word the speaker is currently uttering to update probabilities linking the sound produced by the speaker to phonemes in the native language repertoire of the listener. This, on average, improves the recognition of later words. This model is neutral regarding the nature of the representations it uses (motor vs. auditory). It serve as a reference point for the discussion in Part 3, which proposes a dual-stream neuro-linguistic architecture to revisits claims for and against the motor theory of speech perception and the relevance of mirror neurons, and extracts some implications for the reframing of the motor theory.

## Introduction

Mirror neurons are neurons that are active not only during the execution of a specific, limited class of actions but also during the observation of a related class of actions when performed by others. This paper complements two series of papers: (i) those which ascribe a role for the mirror system for grasping in the evolution of the human language-ready brain, suggesting how mirror neurons may cooperate with systems “beyond the mirror” to play a role in neurolinguistics (Arbib 2006, 2008), and (ii) others developing a perceptuo-motor theory of speech perception known as the Perception-for-Action-Control Theory (PACT; Schwartz et al. 2010).

The *motor theory of speech perception* holds that we recognize the sounds of speech by creating a motor representation of how those sounds would be produced. In earlier papers, the motor theory has been *revised* (Liberman and Mattingly 1985) and *reviewed* (Galantucci et al. 2006) positively, and *revisited* negatively (Massaro and Chen 2008). On the other hand, *mirror neurons* were first discovered as cells correlated with grasps of various kinds in area F5 of macaque premotor cortex (Rizzolatti et al. 1996). This area also contains *canonical neurons* that also fire when the monkey acts but not when it observes the actions of others. F5 contains many other types of neurons as well. Mirror neurons have also been found in monkey parietal cortex (Gallese et al. 2002). Extrapolating this to language, based on data showing that macaque F5 is homologous to human Broca’s area (a key area in the human brain for both spoken and signed language) and that the latter is active when humans both execute and observe hand movements, Rizzolatti & Arbib (1998) suggested as part of the initial formulation of the *Mirror System Hypothesis* (MSH) for the evolution of the human brain’s readiness for language, that “mirror neurons represent the link between sender and receiver that Liberman postulated....” We will dispute this claim as we *reframe* the motor theory (and it is not part of the latest version of MSH, Arbib 2012).

The mirror neurons for manual actions in the macaque include a subset, the *audiovisual* mirror neurons, whose firing is correlated not only with performance and visual observation of a certain type of manual action which has a distinctive sound (e.g., tearing paper, breaking a peanut) but also with hearing the action (Kohler et al. 2002), leading Rizzolatti & Craighero (2004) to suggest that such neurons mediate speech perception. The latter statement needs to be hedged. The audiovisual mirror neurons of Kohler et al. (2002) respond to the sound of a manual action but the sound is not that of the hand *per se* but rather that of the way the hand distorts an object (the paper or peanut, for example). We do know (Ferrari et al. 2003) that the macaque also has "mouth mirror neurons" in premotor area F5 that are active during the execution and observation of mouth actions related to ingestion; and, for a subpopulation of mouth mirror neurons, the most effective visual stimulus is a communicative mouth gesture (e.g. lip smacking), with some also firing when the monkey makes communicative gestures. However, it must be stressed that the control of oro-facial gestures in monkeys does not extend to the concurrent control of phonation: monkey vocal production is limited to a small innate repertoire.

The just mentioned actions with peanuts and paper, as well as other studies with tools (Umiltà et al. 2008; Ferrari et al. 2005), make clear that the mirror neuron repertoire is open to learning. We have modeled the learning processes involved in coming to recognize that an other is performing an action within one's own repertoire (Bonaiuto et al. 2007; Oztop and Arbib 2002) as well as mechanisms which extend one's own repertoire by trial and error (Oztop et al. 2004).

If we assume that the common ancestor of humans and macaques had a macaque-like mirror system for manual and orofacial actions, then it still requires a long evolutionary path to get from such a system to a human brain which extends the macaque-like circuitry in two ways to develop:

(i) a rich system of vocal control which can produce the vowels, consonants and other sounds of modern human languages and a mirror system which can recognize these sounds; and

(ii) a hierarchical system which can extend the recognition and production of basic sounds to ever larger structures such as words, phrases and sentences.

Some authors suggest that the control of vocal musculature came first and provided the basis for spoken language (MacNeilage 1998; MacNeilage and Davis 2005) while others argue that a gestural protolanguage drove the evolution of improved vocal control, with speech eventually dominating over sign thereafter (Arbib 2005). Here we need not depend on a resolution of this debate. Instead, we *assume* that humans (as distinct from monkeys) have mirror neurons for the articulation of speech sounds and address two questions:

(a) Is the motor theory of speech perception correct?

(b) Do mirror neurons play an important role in speech perception?

Lotto et al. (2009) have recently argued that the answer to both questions is "no" – and we will return to their arguments in Part 3 – but we offer the more nuanced answer that

(a) Speech perception need not involve recognition of motor gestures but

(b) The role of the mirror system and gestural recognition becomes increasingly important when noise or neural impairment distorts the speech signal – and also predominates when hearing a non-word (such as a novel family name) that respects the phonology of the language.

Our answer is informed by a computational model of what might seem a clear counter-example to the theory: our ability to recognize our own language when spoken with a foreign accent. The key observation is this: When we meet someone with a novel accent, we may at first understand almost nothing they say, yet after attending closely to a few sentences, we may thereafter follow their speech with relatively little effort.

The paper has three parts:

Part 1 places the motor theory of speech perception in a larger framework, by building on a review of the Perception for Action Control Theory (PACT, Schwartz et al. 2010; Schwartz et al. 2007) and earlier models of the adaptive formation of mirror neurons for grasping. This allows viewing extensions of that mirror system as part of a larger system for neuro-linguistic processing, augmented by the present consideration of recognizing speech in a novel accent.

Part 2 offers a novel probabilistic computational model of phoneme-based word recognition specifically structured to show how a listener comes to understand the speech of someone speaking the listener's native language with a "foreign" accent. We show how this model is able to significantly improve recognition of the speech of a foreign speaker using a relatively small training set. While involving a number of simplifications (which we discuss), we suggest that the core tenet of the model is a reliable approximation: the listener uses hypotheses about the word the speaker is currently uttering to update probabilities linking the sound produced by the speaker to phonemes in the native language repertoire of the listener. We reiterate that this task seems to fly directly in the face of the basic hypothesis of the motor theory, namely that we perceive the speech of another in terms of a motor representation of that speech, more specifically as a sequence of articulatory gestures. Indeed, in motor terms, one might say that the success of the learning is because either

(i) *contra the motor theory*, each candidate word uttered by the speaker is interpreted as a string of abstract phonemes which suffice to support interpretation of a word in the *speaker's* accent even though it has not been heard before; or

(ii) *somewhat motor-theoretic*, but nonetheless a departure from the classic formulation, each candidate word uttered by the speaker is interpreted as the string of articulatory gestures which the *listener* would use to produce the word in the listener's own accent.

In developing the model, we will take an agnostic stand, where the listener builds a "conversion table" from speaker's sound to listener's phonemes, without settling whether those phonemes are auditory, abstract, or motoric in nature. This allows us to emphasize the general issues such a model has to solve.

Part 3 finally returns us to our critique of the motor theory, building on the analysis from Part 1 and the computational model from Part 2. This leads to a novel neuro-linguistic architecture integrating motor and auditory processes. The notion of a dual stream architecture for speech recognition is already well-established in the speech perception literature (Hickok 2009; Hickok and Poeppel 2004) and grounded by studies of dual streams for auditory processing (Rauschecker 1998; Rauschecker and Tian 2000). We emphasize that our new model is compatible with this effort, but is enriched computationally by its roots in the evolutionary extension of a model for grasping (Arbib 2006) informed by re-assessment of studies of dual streams for visual processing (Ungerleider and Mishkin 1982; Goodale and Milner 1992).

# Part 1. Placing the Motor Theory in a Larger Framework

We start with a careful formulation of the phone vs. phoneme distinction. Then we frame our discussion of the status of the motor theory of speech perception by placing it in a larger framework based on a recent review of the Perception for Action Control Theory (PACT), a perceptuo-motor theory of speech perception developed by the Grenoble group of which Moulin-Frier was a member (Schwartz et al. 2010), and an analysis of mirror system activity for action and language (Arbib 2010).

### Phones and Phonemes

*Phones* are the distinctive sounds used by a speaker, whereas *phonemes* are "distinctive" units of sound: /p/ and /b/ are different phonemes because "pin" and "bin" mean different things.[1] However, /p/ can be pronounced in different ways – e.g., /p/ is aspirated in "pin" but not in "spin". Phonologists represent phonemes by writing them between two slashes as in /p/, whereas the actual sounds are enclosed by square brackets. If two similar sounds belong to the same underlying phoneme, they are called *allophones*. For instance, /p/ is aspirated at the beginning of a word in English, but not after /s/ – we thus distinguish aspirated [$p^h$] from unaspirated [p]. In English, /l/ and /r/ are two separate phonemes ("loot" versus "root") but in Korean they are allophones – [r] comes before a vowel, and [l] does not (contrast "Seoul" and "Korea"). A native Korean speaker recognizes the underlying phoneme /l/, and, expresses it as either [r] or [l] depending on the phonetic context. Another Korean will hear both sounds as the underlying phoneme and think of them as the same sound.

We use English as the standard language for the present paper – it will be clear that nothing in our argument rests on this choice. We stress that being a native speaker of English does not imply the use of a standard pronunciation. In Yorkshire, the pronunciation of *blood* rhymes with the last syllable of how one author (MAA) says *understood*; an Australian may say "today" in a way that sounds to others like "to die" /tudaɪ/. Thus, even for native speakers of English there is no unique mapping of words to strings of phones. Going even further, it is not clear when two native speakers with different accents are using the same phonemes "beneath" the phones they employ to pronounce a given word. For example, when the Australian and the speaker of "BBC English" pronounce "today", the underlying phonemes may be the same for each speaker. However, when someone from Yorkshire says "blood" versus "dug" he may be employing distinct phonemes for the vowels, where the speaker of BBC English is using the same vowel. In general, then, the mapping of phonemes as well as phones between two accents will be probabilistic.

We thus see that the distinction between phones and phonemes is central for the problem of foreign-accented speech adaptation, in the sense that the task of the native listener is to map the unusual phones produced by the foreign speaker with her[2] native set of phonemes.

### Perception for Action Control Theory

In communication, certain trade-offs have to be made by the speaker between the effort exerted in producing the signal and the risk of being misunderstood. This is referred to as Hyper-Hypo, or H&H theory (Lindblom 1990): speech production is adaptive and the continuum ranges from hypo- to hyper-articulated speech, depending on the conditions of communication. As evidence that speech perception is not purely motor driven, Perception for Action Control Theory (Schwartz et al. 2010) builds on the observation that the phones used by a speech community are in great part selected for their perceptual value. Take [a i u], the basic vowel system existing in most languages in the world. These vowels are well differentiated along articulatory dimensions, e.g. [a] is low and [i] and [u] are both high, tongue is front for [i] and back for [u]) while lips are unrounded for [i] and rounded for [u]. But consider another vowel system which is equally well-differentiated in terms of articulatory gestures – the system in which tongue and lip configurations are exchanged within high vowels, yielding [y] (front tongue and rounded lips) and [ɯ] (back tongue and unrounded lips). Why is it that *no* human language uses this system unless it is embedded in a very large number of vowels? Not for articulatory reasons – indeed, these elements are also the most motorically distinct – but rather because [y] and [ɯ] are perceptually close, whereas [i] and [u] are very far apart perceptually since lip rounding and tongue backing are synergetic in terms of resonance tuning. Another example in the same vein is given by [d] and [z]: They are produced with a high tongue tip with very close positions. But this small difference in tongue positions results in a dramatic sound change which is great for building phonological systems and languages

[1] See, for example, the article http://www.nationmaster.com/encyclopedia/Phonology

(and, indeed, Whalen et al. 2009 argue for the dominance of tongue contribution over the jaw's, even for vowels distinguished by height). Thus, differentiation for the ear is crucial to the emergence of a phonological system for any language. For that matter, Schwartz et al. (2007) point out that motor theories do not provide any prediction about phonological systems in human languages (and a crucial role of acoustic differentiation in phonological system emergence is also acknowledged by motor production theorists themselves, see Studdert-Kennedy and Goldstein 2003).

Nonetheless, speech perception can well involve motor knowledge though it is not necessary under all circumstances. Much evidence is provided by those who hold the motor theory (e.g., in Galantucci et al. 2006). PACT proposes two basic hypotheses for the roles of the perceptuo-motor link in speech perception:

1) *Developmental:* Schwartz et al. (2010) present data showing that the way a subject produces vowels is related to the way she perceives them. This is related to coarticulation. Initially, an infant or a young child can perceive and distinguish syllables, say [ba] or [bi], purely from the sound (relative formants), which leads to separate classes for each syllable, say [ba] or [bi]. Then the child begins to produce these syllables, tuning her vocal tract to reproduce the sounds of [ba] or [bi]. There the child may discover (subconsciously) that [ba] and [bi] have something in common. This is the birth of phonemes, which can reorganize the perceptual space in a principled way (see blog entries moderated by Hickok and Poeppel 2009). Notice that the argument that phonemes are defined by a fixed set of oppositions is difficult to maintain, since no easy acoustic cue enables one to characterize phonemes or phonemic oppositions for, e.g., labials vs. dentals vs. velars. Rather, we suggest here that perceptual distinguishability may drive processes that yield motor commonalities that are somewhat obscured by coarticulation.

2) *Combating Noise:* In noise, motor simulation can play a key role in organizing the speech scene. PACT considers the perceptuo-motor link as basic in complementing the perceptual representation in case of noise. Although motoric encoding may play little role when comprehending clearly enunciated speech, it may be invoked when we seek to catch an unclearly heard word.

---

[2] "She" and "her" will stand in for "he or she" and "his or her" respectively, unless the context makes clear which gender is intended.

We next complement PACT by offering a deeper understanding of mirror neurons based on computational models of their role in the visual control of hand movements. We will then explore the implications of this for a model of language which allows us to assess critiques of the motor theory of speech perception.

## The Mirror System Hypothesis

We now turn to the mirror system hypothesis on the evolution of the language-ready brain whose earliest version (Rizzolatti and Arbib 1998) was more closely linked to the motor theory than now seems appropriate. To justify this change of perspective we first need to review a model published earlier in *Biological Cybernetics*. Oztop & Arbib (2002) developed the MNS Model of the Mirror System to show how appropriately located neurons in macaque premotor area F5 could *become* mirror neurons through learning perceptual correlates of an executed movement. There were two key inputs. The "perceptual stream" was supplied by parietal input from area PF that combined hand motion data from the superior temporal sulcus (STS) with affordance information to encode the motion of the hand relative to the object. The "motor reference/training signal" was supplied by the activity of F5 canonical neurons controlling actions in the subject's repertoire. By repeated observation of various self-generated "hand state" (hand relative to object) trajectories for a given input from canonical neurons encoding a specific manual action, certain neurons would learn to become mirror neurons for that action, firing not only when activated by the canonical neurons for a specific grasp (self actions) but also by input from PF alone that encoded related hand-state trajectories, whether self-generated or produced by an other. As learning progressed, these mirror neurons learned to recognize "early on" the trajectory relating hand to object during the related actions.

The MNS Model of the Mirror System therefore implies that *sensory data from the same receptor array are processed differentially along different pathways, yielding complementary representations which can nonetheless modulate each other*.

Bonaiuto, Rosta & Arbib (2007) built on this work to develop the MNS2 model, but the above implication of the MNS model is all we need for this paper.

But what of the data, discussed in the introduction, that the macaque has "mouth mirror neurons" and that there is a subpopulation of these for which the most effective visual stimulus is a communicative mouth gesture (e.g. lip smacking), with some also firing when the monkey itself makes communicative gestures (Ferrari et al. 2003)? We have not modeled these, but we can see how an MNS-style model would generate them. We focus on humans, looking at the formation of mirror neurons for facial movements generally. These may be relatively innate, as for some basic emotional expressions and ingestive actions. There appears to be two different issues. One is the so-called *correspondence problem* – matching body part of the self to body part of the other.[3] – and this seems part of the innate machinery of the infant, as demonstrated by the phenomenon of neonatal imitation (Meltzoff and Moore 1977) – and see Ferrari et al. (2006) for data on neonatal imitation in monkeys. The other is learning to recognize the pairing between one's own actions with similar actions performed by others. In some cases there may be auditory cues (e.g., the sound of crying or laughter) or visual cues (the approach of food toward the face). But where the "classical" mirror system can exploit a visual representation of how hands approach objects during both self-action and other-action, an orofacial mirror system cannot access direct visual input on how one's face is moving during self-action. Human infants, unlike monkeys, do have an additional set of cues. Caregivers will often imitate the child's facial expression, and often do so in an exaggerated way. This "augments the database" and gives the child even more cues that enter into the training of mirror neurons for orofacial actions. Data about this type of mirror neurons thus has two other implications:

- *Perception of a given praxic or communicative event may be possible in several modalities yet be enhanced, especially in noise conditions, when consistent multi-modal information is available.*
- *Depending on context, the movement of another's effectors can be interpreted as having, or not having, communicative intent.*

Without going into the details, we may note that another model (Bonaiuto and Arbib 2010) demonstrates the utility of the mirror system's response to the agent's own actions in motor learning. The key implication is that *the mirror system's activation will have no effect during the successful completion of intended movements, but will play a crucial role when intended actions become impaired.*

---

[3] By contrast, in vision the *correspondence problem* is the challenge of matching features extracted from the two retinas (or from the one retina at different times) that correspond to the same feature in the external

Returning to the issue of language, we disputed in the introduction the assertion by (Rizzolatti and Craighero 2004) that audiovisual neurons mediate speech perception but accepted the *working hypothesis* that humans (as distinct from monkeys) have mirror neurons for the articulation of speech sounds. This is part of the elaboration of the *Mirror System Hypothesis (MSH):* Based on the key data that monkey F5 (with its mirror system for grasping) is homologous to area 44 of human Broca's area and that imaging studies show activation for both grasping and observation of grasping in or near Broca's area, the Hypothesis (Arbib and Rizzolatti 1997; Rizzolatti and Arbib 1998) posits that *the brain mechanisms for language parity evolved atop the mirror system for grasping, rooting speech in communication based on manual gesture*. The defense of this evolutionary hypothesis is outside the scope of this paper. Here our task is to address the significance of the generally agreed-upon hypothesis that the human brain contains mirror neurons related to the actions and sounds of speech. What we explore here is a view of neurolinguistics stressing a mirror system for words-as-phonological-actions (Figure 1, adapted from Arbib 2006, 2010). The hypothesis is that actions and words can be recognized in two ways: in the dorsal path, recognition is of the parameterized action (*how* it is executed) or the articulated word (how it is *pronounced*) whereas in the ventral path it is the nature of the action (*what* is being done) or the word (what it *means* or contributes to the overall meaning of the utterance).

Our modeling (Fagg and Arbib 1998) adds an important point about the relation of the dorsal and ventral streams. While the dorsal stream may compute parameters for each visual affordance of an object, any one of which can be chosen if the task is merely to grasp the object, it requires object recognition and prefrontal processing concerning the task at hand to determine which affordance is best employed for other tasks – and this requires connections from PFC to AIP and/or F5 which can connect the "planning-level part of the object" (encoded in IT and PFC) to the affordance for that part of the object which will enable AIP and F5 to inform primary motor cortex (F1) about the actual parameters needed to, e.g., grasp the handle given the current size of the handle and its location and

---

world.

orientation relative to the actor – parameters crucial for muscle control but not necessary in nearly as much detail for deciding whether to grab the mug and if so whether to grasp the handle and bring the mug to the lips, or grasp the rim to move the mug to a new location.

This view of grasping is congruent with the two-stream model of speech perception offered by Hickok & Poeppel (2004; Hickok 2009), though, as stressed in the introduction, our new framework is rooted in our model of the emergence of the language-ready brain (Arbib 2012; Rizzolatti and Arbib 1998), so that congruence with the work of Hickok and Poeppel provides a test of the theory, not the basis for it. They posit that early cortical stages of speech perception involve auditory fields in the superior temporal gyrus bilaterally (although asymmetrically). This cortical processing system then diverges into

- *a ventral stream which is involved in mapping sound onto meaning*, and projects ventro-laterally toward inferior posterior temporal cortex (posterior middle temporal gyrus) which serves as an interface between sound-based representations of speech in the superior temporal gyrus (again bilaterally) and widely distributed conceptual representations.
- *a dorsal stream which is involved in mapping sound onto articulatory-based representations* and projects dorso-posteriorly involving a region in the posterior Sylvian fissure at the parietal–temporal boundary (area Spt), and ultimately projecting to frontal regions. This network provides a mechanism for the development and maintenance of "parity" between auditory and motor representations of speech.

The notion is that the ventral stream can support the understanding of speech in normal conditions without recourse to the dorsal path, whereas the dorsal stream connects sounds to motor commands and could be involved in low-level imitation or in adverse speech perception. Linking sounds directly to motor commands, the dorsal stream could provide a functional role for mirror neurons. In particular, the dorsal stream would have to take the lead role in recognizing a non-word such as *perfuddle*. More generally, though, the two streams work together. Thus if the speaker generates a sound that is ambiguous between *bin* and *pin*, the matter could be resolved, say either by a stronger activation of neurons representing (features of) /p/ than for /b/, or the competitive activation of *bin* and *pin* might be biased at the word level or above by a context like *you could hear a ___ drop).*

Turning to the perception of actions, we note an fMRI study of humans observing actions performed by both humans and nonconspecifics (Buccino et al. 2004). To simplify somewhat, they observed activity in the frontal area associated with the mirror system when humans observed videos of biting movements, whether those of another human, a monkey or a dog. But when it came to communicative movements of the mouth, observing human lip movements for speech activated the “mirror system”, observation of lip-smacking in a monkey activated only a small focus in the region, bilaterally, while the observation of a dog barking did not produce any activation in the frontal lobe. However, observation of all types of mouth actions induced activation of extrastriate occipital areas.

Buccino et al. hypothesize that actions belonging to the motor repertoire of the observer (e.g., biting and lip movements for speech) are mapped on the observer's motor system whereas actions that do not belong to this repertoire (e.g., barking) are recognized based without such mapping. However, we find it more plausible to argue that *all* recognizable actions have a semantic representation in the ventral path, but that these are reciprocally connected with motor representations when the action is in the observer’s repertoire. Note that although (most) humans do not bark, they can imitate barking – one might thus expect a different result for brain activation for “observe in preparation to imitate” as distinct from mere observation. Similarly, note the difference for motor representations for an action, the pantomime of an action, and the action of speaking a word. Note, further, that most words are not verbs, and a fortiori do not correspond to actions, but speaking, signing or writing the word is an action related either to other semantic content (schemas of the ventral path) or to some grammatical function.

The key data for MSH place a mirror system for grasping actions, and thus (inferentially, as a result of evolution of the brain) for language actions, in Broca’s area, suggesting that the prime effect of damage to Broca’s area would be the inability to pronounce words. However, Broca’s aphasia (which typically involves a lesion that extends well beyond Broca’s area) involves *agrammatism,* a pattern of syntactically defective speech that may range from production only of one-word utterances to mildly 'telegraphic' speech. Thus, the transition from *Actions* to *Compound Actions* and from *Words* to *Constructions* provides key challenges for research. A key part of the Mirror System Hypothesis is that the mirror system evolved in human ancestors as part of successively larger, more competent systems: an enlarged system to support *simple imitation*, the ability to acquire some novel actions by extensive observation and repetition, but only on a limited basis (this ability goes back to the common ancestor of humans and apes); thereafter, *complex imitation* evolved in the human line since the divergence from the great apes, with imitation based on the ability to observe a novel performance and see, to a first approximation, its key subgoals and the actions which appear to achieve them. The Mirror System Hypothesis posits that complex imitation evolved to support praxic actions, but then contributed to the evolution of communication by making possible the *lifting from words to constructions and from phones to words* that distinguishes language from other forms of communication**.**

The next part proposes a computational model of foreign-accented speech adaptation which is neutral as to whether or not phone representations are motor, auditory or abstract. Part 3 will then assess its relevance to the reframing of the motor theory of speech perception.

# Part 2. Modeling Foreign-Accented Speech Adaptation

Our aim in this second part is both:

- to review related research on perceptual learning and adaptation in speech in order to highlight necessary conditions in terms of cognitive processes needed to solve the problem of foreign-accented speech adaptation,
- to provide a simple computational model implementing these processes, allowing us to show that these conditions are also sufficient to fit the experimental data.

Thus, this latter must be viewed as a neutral platform which will emphasize the general issues such a model have to solve, that we will then rely in Part 3 to the processes involved in the neurolinguistics analysis we propose in Part 1

## Related Research on Perceptual Learning and Adaptation

Perceptual learning consists of extracting previously unused information from the environment to perform specific tasks and involves relatively long-lasting changes to an organism's perceptual (and, in many cases, motor) system, whereas perceptual adaptation rather refer to transient effects of unusual stimuli (Goldstone 1998). We thus consider our problem of foreign accent perception as a specific instance of perceptual adaptation. We report here some related work in the more general domain of speech perception.

Our aim is to identify key points to solve the problem of rapid foreign-accented speech adaptation, in particular:

- Learning: how to use lexical feedback to update the perceptual knowledge of the listener; and how new information can be integrated into previously learnt perceptual knowledge?
- Generalization: rapidly learning how to make generalizations in order to improve recognition rate on novel stimuli.
- Categorization: Is sub-lexical categorization necessary for speech recognition, and/or for learning generalization?

Bradlow and Bent (2008) provide a study on "Perceptual adaptation to non-native speech" which confirmed our key observation that when we meet someone with a novel accent, we may at first understand almost nothing they say, yet after attending closely to a few sentences, we may thereafter follow their speech with relatively little effort. Their experiment consists of a set of sentences pronounced in English by speakers with a foreign accent (the Northwestern University Foreign-Accented English Speech Database, NUFAESD) and presented to native English listeners who were then asked to transcribe the heard sentences in standard English orthography.

For each speaker, the database includes an intelligibility score, measured independently of the experiment as a recognition rate, which defines a talker's baseline level of English sentence intelligibility.

The sentence stimulus is divided into four quartiles. This allows analysis of how the recognition rate of the listeners evolves with the time of exposure to a given foreign-accented speech stimulus.

The authors found that the amount of exposure required in order to achieve significant improvement in intelligibility increases as baseline intelligibility decreased, which is consistent with the view of an integration of information across levels of representation, and therefore to some extent to the approximation we made in the introduction: the listener uses hypotheses about the word the speaker is currently uttering to update probabilities linking the sound produced by the speaker to phonemes in the native language repertoire of the listener. In the same spirit, Norris et al. (2003) demonstrates that listeners use lexical knowledge in perceptual learning of speech sounds, Eisner and McQueen (2005) extend this result by showing that perceptual adjustment can be highly specific both with respect to segmental information and with respect to information about talker identity.

The interaction between levels of representation is also described in the Adaptive Resonance Theory (ART, Grossberg 2003), which suggests that conscious (speech) percepts arise from a resonance between bottom-up signals activated by environmental events and top-down expectation learned from previous experience. ART is a mechanism to control rapid learning of perceptual and cognitive code without catastrophic forgetting.

This view of perception as a resonant state between signals and memories is also present in Goldinger (1998), who suggests an episodic memory of lexical access, where words are perceived against a background of myriad, detailed episodes. For this aim, he applies the episodic memory model MINERVA 2 (Hintzman 1986) to speech perception and tests it against speech production data from a word-shadowing task. This “episodic theory” contrasts with standard “normalization theories”, where idiosyncratic aspects of speech (voice details, ambient noise, etc.) are considered noise and are filtered in perception.

According to Pierrehumbert (2002), such an episodic (or exemplar) theory involves a map of the perceptual space and a set of labels over this map. For phonetics, the perceptual map can involve acoustic as well as articulatory features and the labels over the map are the inventory of phonological primitives. Thus, each label is associated with a large set of remembered percepts, although the latter can be summarized in an updatable frequency distribution over a cognitive map.

Klatt (1979) designed a model of lexical access from spectra, LAFS, based on spectral sequence decoding structure, in which phonetic segments and phonological rules play an important role in network compilation, but not in the direct analysis of the speech waveform during lexical search. The paper also describes a model of phonetic transcription from spectra based on a similar modeling technique, SCRIBER, but which is independent of LAFS. This separation of lexical access and phonetic decomposition into two independent systems is driven by the belief that lexical hypotheses can be generated rapidly and more accurately in a LAFS structure, where knowledge is pre-compiled, than in any two-step model (phonetic recognition followed by lexical access) containing the same acoustic-phonetic and phonological knowledge. However, phonetic recognition seems to be necessary to learn unfamiliar words, as it is proposed by the author as an extension, interfacing LAFS and SCRIBER with higher-level cognitive structures (top-down lexical predictions).

These latter views (interaction between levels of representation, episodic memory of lexical access, independence between lexical access and phonological decomposition) are well integrated in the Polysp conceptual framework (polysystemic speech perception, Hawkins 2003), which suggests that fine phonetic detail, stored as multi-modal exemplar memories linked to non-linguistic and linguistic information, plays a crucial role in how people understand ordinary conversational speech. Polysp assumes that "*there is no one way to understand a speech signal: the polysystemic linguistic structure can be identified by many routes, in different order or in parallel*". Thus, the meaning access of a speech signal does not necessarily involve a clear identification of every sub-units it contains: given the context, understanding can happen by recognizing only a few parts of the signal, and a particular phonetic realization can contribute to the meaning of an utterance (e.g., saying *"dunno"* in a given context can contain phonetic information enough to understand that the speaker do not know, and can also convey the meaning that the speaker is indifferent to the listener's wish of information).

Following the assumption of exemplar memories in speech, Pierrehumbert (2002) suggests that speech production can be biased by perceptual experience with specific words, thus assuming an interaction between speech production and perception.

The neuro-computational model of Guenther et al. (2006) implements a Speech Sound Map (typically, a phoneme string) which is linked to motor, somatosensory and auditory representations. Two paths collaborate in speech production: a feed-forward system directly linked to motor commands and a feedback system modulating the motor commands from auditory and somatosensory error signals. (See also Kröger et al. 2009.)

Francis et al. (2000), Francis and Nusbaum (2002), as well as Best et al. (2001) studied the acquisition of novel phonetic categories in the case of a second language. Although the problem is not exactly the same than the adaptation of foreign-accented speech in the own language of the listener, these studies provide materials to the questions of learning, generalization and categorization: what is learnt? How it is generalized to novel stimuli? How it is integrated with the previously learnt perceptual knowledge by merging or splitting existing phonological categories? But the answers to these questions are still discussed and our choices in this respect will be made clear in our problem formalization.
This review thus suggests that a model which aims at render an account of foreign-accented speech adaptation should integrate the following points:

1. A two-step model (sub-lexical categorization followed by lexical access) is not necessarily involved, or even useful, in normal speech perception (see the LAFS vs. SCRIBER distinction, as well as the role of phonetic details).
2. During perceptual adaptation to foreign-accented speech, the listener must integrate the information provided by the novel stimuli with her existing perceptual knowledge on her native language.
3. To correctly adapt her perception, the listener needs a training signal (a feedback), and lexical hypotheses about easily recognizable words seem plausible.
4. In order to generalize the learning, the listener has to learn at a sub-level of the feedback, that is at a sub-lexical level according to the previous point (otherwise she will be unable to generalize her learning to novel words, and thus will be unable to rapidly adapt her perception).
5. As a corollary to the previous point, the listener must be able to make categorization at a sub-lexical level to generalize her learning, although such a categorization is not necessary for normal speech recognition (see the LAFS vs SCRIBER distinction in Klatt 1979).

## Hidden Markov Models

As we just saw, numerous works show that perceptual learning and adaptation arise in speech processing, and that it can be lexically driven (Norris et al. 2003; Eisner and McQueen 2005). However, we still do not know how variability is handled in speech perception. Computational model of speech adaptation are therefore useful, in the sense that they can emphasize which processes are necessary or sufficient for this kind of tasks. According to Eisner and McQueen (2005), no current model of word recognition can accommodate perceptual learning at a segmental level.

The use of Hidden Markov Models (HMMs) to map acoustics to lexical forms is standard commercial practice, giving the kind of plasticity that yields speaker adaptation and the means to augment the system with new phonetic information (Gales and Young 2007). Basically, a HMM can be conceived as a probabilistic finite state machine, which is able to emit observations during state transitions. In the case of automatic speech recognition applications, those states generally represent sub-lexical categories, whereas observations represent the acoustic input. The states are “hidden” in the sense that only the acoustic input is available to perform lexical access.

HMMs are an appropriate tool to solve the problem of foreign-accented speech adaptation because they provide a unified formalism to:

- compute the probability that a given HMM (representing a lexical unit, a word for example) emits a given sequence of observations (the auditory input), by exploring all the possible state sequences (i.e. sub-lexical unit sequences) able to produce that observation sequence, that is without being a two-step model where lexical access is preceded by sub-lexical categorization (direct lexical access, in the spirit of LAFS, point 1 above);
- given an HMM and an observation sequence, to find the most probable state sequence the observations come from (sub-lexical decomposition, in the spirit of SCRIBER, point 5 above);
- given an HMM, an observation sequence, and current parameters of the model, to refine the parameters to fit better with the observations (adaptation to novel stimuli using lexical hypotheses, point 2, 3, 4 above).

These characteristics of HMMs make them good candidates to solve the key points of our problem regarding learning, generalization and categorization.

We thus offer an HMM model that addresses a specific task: Given that the model can already recognize words when they are pronounced in accent N, to have the ability to “bootstrap” recognition of words in a new accent F, gaining fast access to a reasonably accurate level of recognition prior to a lengthy process of improving word-by-word recognition.

As we argued in the review of the literature, a sub-lexical categorization seems to be necessary to generalize the learning to novel words. Thus, where the expert phonetician has a “universal” set of symbols that cover the phonemes used by speakers of many different languages and diverse accents, our native speaker N starts with a limited set which may not include all those in F’s repertoire. To formalize N’s increasing mastery of F’s accent, we need to characterize what N hears before F’s accent is mastered. To simplify the problem, we assume (i) that N has no trouble separating F’s speech into separate “word candidates” but at first may not recognize what those words are; and (ii) that N hears each word as a string not of phonemes but rather of phones, represented as feature vectors, where the features are rich enough to encode both N’s and F’s stock of phonemes (and many more besides). We thus leave aside the issue of how N can attend to the auditory input in such a way as to discriminate these features (an effort in attention may certainly be required until the accent is mastered).

Yet most HMM-based modeling techniques for automatic speech recognition use acoustic signal as input (often preceded by feature extraction), word (sequence) as outputs, and phones as hidden states (Gales and Young 2007; Adda-Decker 2001), the theoretical notion of phoneme being generally inappropriate for a performance-driven system. But we want to make clear that our aim here is not to design a technologically-competitive automatic speech recognition (ASR) system, nor to advance the state of the art in the domain: the results that we will show in this part will not be surprising to the ASR community. Instead, our point is theoretical: we want to design a simple model implementing the key points we identified above about learning, generalization and categorization, and to show that it is sufficient to perform foreign-accented speech adaptation in a reasonable way, and finally that it supports our analysis in Part 1. Consequently, we will make non-standard choices in our HMM-based modeling which are more appropriate to our specific problem of foreign-accented speech adaptation where the phones vs phonemes distinction is central: inputs will be phones (represented as feature vectors), outputs will be words, and hidden states will be phonemes of the native language.

In the motor theory (Liberman and Mattingly 1985; Liberman and Whalen 2000), or in Articulatory Phonology more generally (e.g., Goldstein et al. 2006), each unit (let's say a phoneme) is represented by the articulatory gesture which produces it. In the present part of the paper, we adopt a neutral stance, trusting the reader to see that the learning method we offer applies to any phonetic representation, whether based more on sound or on action. This provides a neutral platform for our discussion of the motor theory and mirror neurons in Part 1. In short, the model we are going to describe is not, nor is it intended to be, a test of the Motor Theory (although such models comparing motor- and auditory-based speech recognition in various conditions exist, see for example (Kirchhoff 1998), or (Moulin-Frier et al. 2012)). Rather, our model provides a conceptual reference point for the theoretical reframing of the motor theory we will offer in Part 3.

## Feature space

As a basis for defining the space $\mathcal{V}$ of features with which to characterize known and novel phones in our model, we turn now to the representation of the International Phonetic Association (1999) for vowels (Figure 2) and consonants (Figure 3). It will be clear that our model can be applied to representations of words which employ different features, such as clicks or tones, or even signed languages, but these extensions are not germane to the present study.

Note that these symbols can represent both phones and phonemes: [ɹ] and [l] for example, are two different phonemes in English, but are allophones of the same phoneme (say /l/) in Korean.

For a phone p, $D_i(p)$ denotes the $i^{th}$ dimension of the feature vector of p. The first dimension $D_1$ of those vectors indicates the type of the phone: V = vowel; C = consonant. The definition of the other dimensions differs with the value of the first dimension as V or C:

**The definition of vowel features** is based on Figure 2, with each range defined on integer values. If v is a vowel ($D_1(v) = V$) then:

$D_2(v)$ is the Open/Close dimension and ranges from 1 (close) to 7 (open).

$D_3(v)$ is the Back/Front dimension and ranges from 1 (back) to 3 (front).

$D_4(v)$ is the Rounded dimension and ranges from 0 (not rounded) to 1 (rounded).

$D_5(v)$ is set to 0 for a non-diphthong and 1 for a diphthong. A diphthong is a unitary vowel with a smooth movement of the tongue from an articulatory point to another in a less open configuration, such as can be realized in one closing gesture (for example /aʊ/ in /saʊnd/ ("sound")). For the sake of simplicity in our modeling, we focus on the fact that a diphthong is perceptually close to the vowel defined by the initial articulatory configuration, and we limit the influence of the final one to its rounding value. In this way, a diphthong is defined as the merge of two consecutive vowels $v_1$ and $v_2$, where $D_2(v_2) \leq D_2(v_1)$, (the initial articulation point is more open than the final one), into a single vowel v identical to $v_1$ except the value of $D_5$ which is set to 1 if $D_4(v_1) = D_4(v_2)$ ($v_1$ and $v_2$ have the same rounding value), 2 if $D_4(v_1) \neq D_4(v_2)$ ($v_1$ and $v_2$ do not have the same rounding value). The vowel $v_2$ is then removed. A diphthong is thus considered to be further from $v_1$ if the rounding value of $v_2$ is different than if it is not.

$D_6(v)$ is the Nasal dimension and takes values 0 (not nasal) or 1 (nasal).

**The definition of consonant features** is based on Figure 3. Again, each range is defined on integer values. The two dimensions of the table in Figure 3 are represented by $D_2$ and $D_3$ except for nasality which is instead represented in $D_4$ with the voice information. If c is a consonant phone ($D_1(c) = C$):

$D_2(c)$ is the Place dimension, defined in the range 1 (glottal) to 11 (bilabial).

$D_3(c)$ is the Manner dimension, defined in the range 1 (lateral approximant) to 7 (plosive). The Nasal value of $D_3$ which appears in Figure 3 is not represented here but in dimension $D_4$.

$D_4(c)$ is the Voice dimension, taking value 2 for a non-voiced consonant, 1 for a voiced one, and 0 for a nasal one. We chose this representation because there is an increasing intensity of the signal from a non-voiced to a voiced to a nasal consonant.

$D_5(c)$ represents affricate consonants (consonants which begin as plosive and release as fricative in the same place, as 'ch' in 'cheese' or 'j' in 'jeans'). An affricate is coded as the corresponding fricative, except for $D_5$ which takes the value 1 (instead of 0). For example, the affricate tʃ ("**ch**eese") is coded as ʃ ("**sh**e") but with $D_5 = 1$.

Given our motivating concern with the motor theory of speech perception, it is striking that the phones, are defined in Figures 2 and 3 primarily by a vector of *motor* features, though when a phonetician transcribes speech she will generally base the transcription on the sound of the phone, lacking direct information on the motion of the articulatory apparatus. Of course, there is a statistical element here in that a given sound pattern ϕ may sound most like a "standard" /f/ or somewhat more like a /v/, so that a transcription "collapses" the original speech signal, but this is not unlike our categorical speech perception in which we tend to reduce what we hear to the phones with which we are familiar.

For each speaker there is a neural motor program for producing a certain set of phones in the sense that, when that person is speaking slowly and clearly, a phonetician's reading of a transcript would coincide with the speaker's own understanding of what was just said – this despite the fact that a given speaker will produce somewhat different spectrograms in producing the same string of phones on different occasions. A phonetic transcription generally combines the phonetic symbols with diacritics which represent additional features, for example the duration of a vowel. In this study, we do not take these into account, except for vowel nasalization (represented with a tilde symbol above the corresponding phonetic symbol in the phonetic transcriptions).

### Corpus

The data for testing our model are provided by the Speech Accent Archive (Weinberger 2010), a corpus of phonetic transcriptions of the same English elicitation paragraph spoken by speakers with many different native languages. Here is the elicitation paragraph:

*Please call Stella. Ask her to bring these things with her from the store: Six spoons of fresh snow peas, five thick slabs of blue cheese, and maybe a snack for her brother Bob. We also need a small plastic snake and a big toy frog for the kids. She can scoop these things into three red bags, and we will go meet her Wednesday at the train station.*

Without loss of generality, we take N to be a speaker whose accent corresponds to the MRC Psycholinguistic Database (Wilson 1988) which provides the Brown verbal frequency (Brown 1984)[4] and the phonetic transcription of around 10,000 English words. The inventory for N comprises 2000 words: the 55 different words of the elicitation paragraph and the remaining 1945 words of the MRC Psycholinguistic Database with the highest frequency. Note that, except to build the native inventory, word frequencies are not taken into account in the model.

---

[4] The Brown verbal frequency is the frequency of occurrence in verbal language derived from the London-Lund Corpus of English Conversation by Brown (1984).

We chose transcriptions from classical dictionaries for the few words of the elicitation paragraph which are not in the MRC database (see Table 2).

The transcription of the elicitation paragraph for N is then:

[pliz kɔl stɛlə æsk hɜ tu bɹɪŋ ðiz θɪŋz wɪð hɜ fɹɒm ðə stɔ sɪks spunz ɒv fɹɛʃ snəʊ piz faɪv θɪk slæbz ɒv blu ʧiz ænd meɪbi ə snæk fɔ hɜ bɹʌðə bɒb wi ɔlsəʊ nid ə smɔl plæstɪk sneɪk ænd ə bɪg tɔɪ fɹɒg fɔ ðə kɪdz ʃi kæn skup ðiz θɪŋz ɪntʊ θɹi ɹɛd bægz ænd wi wɪl gəʊ mit hɜ wɛnzdɪ æt ðə tɹeɪn steɪʃən]

We may compare this with the phonetic transcription for a French speaker of English (French8):

[pliz kɔl stɛlə ɐsk hɜ tə bɹĩŋg ði̞s θĩŋgz wɪθ hɜ fɹʌ̃m n̪ə stɔ sɪks spũnz əf fɹɛʃ ʃnoʊ piz faɪfs θɪk slæβ̥s əf blʉ tʃiz ɛ̃n meɪbi ə snɐk fɔ̃ hɜ bɹʌðə bop wi ʌlso nid ə smɔl plæstɪk sneɪk ɛ̃n̊ ə bɪk tɔɪ fɹog fɔ̃ ðə kɪd̥s ʃi̥ kɛ̃n skʉp ðis θĩŋz ĩntu θɹi ɹɛd bægz ɛ̃n wĩ wɪl gɵʊ mit hɜ wɛ̃nzdeɪ æd̪ ðə tɹeĩn steɪʃən]

Our model will process data given in the form of such transcriptions, rather than using, e.g., spectrograms from actual speech patterns. Of course, this makes the problem simpler for the model than for the human, because we posit that N can unequivocally determine the sounds emitted by F. Two points need to be made:

(a) Although we have chosen to characterize these sounds by the bundle of features used by the IPA (Figures 2 and 3), our model is neutral as to what features are detected by N (they could be articulatory gestures). It demonstrates that, whatever features are detected, the foreign-accented speech adaptation problem can be solved.

(b) An expanded model would also address the issue of how to handle uncertainties for N in detecting which features were actually employed by F on a particular occasion.

But why not use actual speech and work with an off-the-shelf HMM-based large-vocabulary continuous speech recognition system? And isn't the significance of results based on hand-coding a dataset containing only 69 words questionable? Our answer is that (despite its shortcomings), the pre-coded database allows us to test the model against a wide range of different accents, using a database that is widely accessible to other researchers. Our concern is not to train an HMM to recognize a vocabulary in a new accent *ab initio,* but rather to show how a process of phonetic substitution, with additions and deletions, can be rapidly acquired as a means to short-circuit the learning process, transferring N's experience with a small set of words pronounced by F to allow fairly accurate recognition of other words spoken by F that N has not heard before. Our aim here is not to advance the state of the art in speech recognition but rather to provide a model adequate to assess the analysis of the motor theory of speech perception in Part 1. The Speech Accent Archive provides an appropriate data set for this focus. The demonstration that we see a consistent effect with only 69 words (but with diverse accents) is a positive one – which is not to deny that it would be an interesting challenge for future work to explore how a variant of the model would perform on any "typical" set of sentences containing a typical vocabulary.

## Formalizing the Problem

To formalize the accent recognition problem, we say that for each speaker S there is a partial map $\Psi_S$ from $\mathcal{P}$ (the set of English phonemes) to $\mathcal{V}$ (a feature space for phones) which maps each phoneme p to the feature vector $\Psi_S(p)$ used by S to pronounce it, and a map $\Phi_S$ from $\mathcal{W}$ (a set of words of English) to $\mathcal{P}^*$ (the set of strings of English phonemes) which maps each word w of English in the speaker's vocabulary to the string of phonemes $\Phi_S(w)$ that S uses to pronounce it.

Extending this notation, $\Psi_S(\Phi_S(w))$ is the sequence of feature vectors that S will produce when saying the word w slowly and clearly. Note that in the present model, we ignore the problem of coarticulation. This would require an extension of the model to learn how the features of a phone vary with its immediate context as well as with long range dependencies such as lip-rounding or nasalization – improvements important in detailed applications to modeling of human data or to technology, but unimportant for our primary goal in this paper, the reframing of the motor theory in Part 1.

The problem we wish to solve then is to provide a statistical method whereby N, by listening to F, can learn to "decode" what F says, despite a difference in English, i.e., to infer a (partial and probabilistic) function $G_{F\rightarrow N}$: $\mathcal{V}^* \rightarrow \mathcal{W}$ from the string of phones that is actually heard (in the "foreign" accent) to the word of English that F intended to say by uttering this string. We thus require that $G_{F\rightarrow N}(\Psi_F(\Phi_F(w))) = w$ with high reliability to reach a suitable recognition rate for the speaker F.

Our claim is that this can be achieved after hearing relatively few sentences and thus a very small subset of $\mathcal{W}$: i.e., a relatively small number of words will enable our model of the listener N to build (implicitly) a table of what each feature vector used by F to pronounce a word may correspond to when N pronounces the word. To achieve this, our hypothesis is that N uses easily recognizable words from F's speech as a training signal (or feedback) to build the table. More formally, we claim that

(i) for easily recognizable word w in some small set $\mathcal{W}_1$, pronounced by F as $\Psi_F(\Phi_F(w))$, N is able to use its initial maps $\Phi_N$ and $\Psi_N$ to approximate $\Phi_F(w)$ and $\{\Psi_F(p) \mid p \in \Phi_F(w)\}$ ; and then

(ii) N uses this to derive a map $H_{F\rightarrow N}$ from $\mathcal{V}$ to $\mathcal{P}^*$ such that for most words in $\mathcal{W}$, if $\Psi_F(\Phi_F(w))$ equals the string of feature vectors $v_1v_2 \ldots v_k$ then $\Phi_N(w)$ equals (more or less) the string of phones $H_{F\rightarrow N}(v_1)$ $H_{F\rightarrow N}(v_2) \ldots$ $H_{F\rightarrow N}(v_k)$. What will make the later model somewhat complicated is that F may not only use different phones (as described by $H_{F\rightarrow N}$) but may also add or drop some phones relative to N's pronunciation of w.

In this model, the native agent N learns the association between its own native phonemes and the feature vectors perceived from F's speech. More formally, using the notations above, the training set is constituted by a set of couples $(\Phi_N(w), \Psi_F(\Phi_F(w)))$, where $\Phi_N(w)$ is the phoneme sequence that N used to pronounce the word w, and $\Psi_F(\Phi_F(w))$ is the sequence of feature vectors that N perceives from F's speech when this latter pronounces w. N then uses this training set to approximate the map $\Psi_F$ and $\Phi_F$.

### Probabilistic modeling

Henceforth, we assume a native English agent N characterized by:

- The map $\Phi_N$ from $\mathcal{W}$ to $\mathcal{P}^*$, which provides the phonetic transcription of each word in the native inventory using the MRC Psycholinguistic Database.
- The map $\Psi_N$, which provide a representation of native phonemes as feature vectors, defined as a conditional probability distribution $P(v \mid p)$, where p is a native phoneme and v is a feature vector.

Using this knowledge, we first propose a word recognition system based on Hidden Markov Models (HMM, Rabiner 1989), where each word is modeled as a single HMM, which competes with each other to find the most probable word for a given heard feature vector sequence. Then we will define the free parameters of the model, corresponding to insertion, deletion and production probabilities, which are shared among HMMs. Finally we present a foreign accent learning process which is able to improve significantly the recognition rate using only a small training set, constituted by a subset of the corpus words associated with F's pronunciation, to tune the parameters.

### *Principles*

The problem for N in recognizing a word pronounced by F can be expressed as follows: given a sequence of feature vectors perceived from speaker F, what is the most probable word in the inventory of N?

F's pronunciation and N's representation of a word w can differ in two ways:

- F's pronunciation can lead to possible insertion, substitution or deletions in $\Phi_N(w)$
- For a given phoneme $p \in \Phi_N(w)$, F can pronounce a different feature vector than N's expectation $\Psi_N(p)$.

Based on the knowledge of $\Phi_N(w)$ and $\Psi_N$, we derive for each word w a Hidden Markov Model (HMM) R(w). The aim is both:

- Word recognition: to find the word w in N's native inventory for which the HMM R(w) is the most likely to produce the feature vector heard from F.
- Accent learning: to find the most likely sequence of insertions, productions, and deletions which allows a given HMM R(w) to produce a given feature vector heard from F, and update the parameters of R(w) accordingly.

If the phonetic transcription of w is $\Phi_N(w) = (p_1,\ldots,p_n)$, the states of R(w) are $S_1$, $S_2$, …, $S_n$, $S_{n+1}$, $S_{n+2}$, where $S_1$ is the unique initial state and $S_{n+2}$ the unique final state. For each state $S_i$ with $1 \le i \le n$, we associate the $i^{th}$ phoneme of w, $phon(S_i) = p_i$. The possible transitions between two states are the following:

- For all $1 \le i \le n+1$, there is a transition from $S_i$ to itself with the unconditional probability P(ins) of inserting a feature vector. Such a transition produces a feature vector v with the probability P(v|ins). Note that both P(ins) and P(v|ins) are independent of the state $S_i$.

- For all $1 \leq i \leq n$, there is a transition from $S_i$ to $S_{i+1}$ with the probability P(del|phon($S_i$)) of removing the phoneme phon($S_i$). Such a transition does not produce any feature vector.
- For all $1 \leq i \leq n$, there is a transition from $S_i$ to $S_{i+1}$ with the probability P(prod|phon($S_i$)) of producing the phoneme phon($S_i$). Such a transition produces a feature vector v with the probability P(v| phon($S_i$)).
- Finally, there is a transition from $S_{n+1}$ to the final state $S_{n+2}$ with the probability 1-P(ins). This transition only aims to terminate the process and does not produce any feature vector.

Comparable models are described by Bahl and Jelinek (1975) and Pinto and Szoke (2008). The following are the free parameters of the model (they will be quantified later):

- P(ins): the unconditional probability of inserting a feature vector,
- P(v|ins): the distribution over $\mathcal{V}$ of inserted feature vectors,
- P(del|p): the probability of deleting a given phoneme p,
- P(prod|p): the probability of producing a given phoneme p,
- P(v| p): the distribution over the feature space $\mathcal{V}$ for a given phoneme p to produce, corresponding to the map $\Psi_N$.

For all state except the final one, the outgoing transition probabilities must sum up to 1. We will see how we ensure this property when we quantify the free parameters.

Figure 4 shows the resulting HMM R("need") for the word "need" with the phonetic transcription $\Phi_N$(need) = /nid/.

Before going further in the algorithmic details of HMM's computation, let us recall the three problems that an HMM is able to solve (according to Rabiner 1989 and adapted to our notations):

- P1: Given an observation sequence in $\mathcal{V}^*$, what is the probability that it comes from a given HMM R(w)?
- P2: Given an observation sequence in $\mathcal{V}^*$ and an HMM R(w), what is the most likely state sequence in $\mathcal{P}^*$ the observations come from?
- P3: Given an observation sequence in $\mathcal{V}^*$ and an HMM R(w), how can the free parameters of the model be refined?

The relation between these three problems are schematized Figure 5.

The following three sections detail the algorithmic implementations of P1, P2 and P3.

*Word recognition (P1)*

Given a sequence of feature vectors $v_1,\ldots,v_m$ perceived by N from the speaker F, the word recognition problem corresponds then to finding the word $w \in \mathcal{W}$ which maximizes the probability that $v_1,\ldots,v_m$ was produced by R(w): $P(R(w)|v_1,\ldots,v_m)$. A key simplification here is to assume that the listener N can indeed interpret the sounds that F is making in terms of available feature vectors.
According to Bayes rule:

$$P\left(R(w)\middle|v_1,\ldots,v_m\right) = \frac{P(R(w))P\left(v_1,\ldots,v_m\middle|R(w)\right)}{P\left(v_1,\ldots,v_m\right)} \qquad (1)$$

Although different words appear in the overall corpus with different frequencies, we assume that in the situation in which N seeks to decode F's accent, these context-independent frequencies are irrelevant, since contextual factors will bias the set of words under consideration. We thus treat each word (and thus their corresponding HMM) as equiprobable in this context, so the term P(R(w)) is taken to be uniform.
In addition, $v_1,\ldots,v_m$ is known when evaluating (1), so the word $w \in \mathcal{W}$ which maximizes $P(R(w)|v_1,\ldots,v_m)$ is therefore the same as that which maximizes $P(v_1,\ldots,v_m \mid R(w))$, the probability that R(w) produces $v_1,\ldots,v_m$. This corresponds to the sum of the probabilities of each individual state sequence from the initial to the final state in R(w) that leads to producing $v_1,\ldots,v_m$. All these possible sequences are represented in the lattice of Figure 6 for an example with m=3 and a 2-phoneme word w. Thus, each possible state sequence leading R(w) to produce $v_1,\ldots,v_m$ corresponds to a path in the lattice derived from R(w). The probability of a path is then the product of the weight of the arrows traversed on it.
For a given word w, let $P(v_1,\ldots,v_k, S_i \mid R(w))$ be the probability that R(w) produced $v_1,\ldots,v_k$ and is in state $S_i$. We need to compute $P(v_1,\ldots,v_m, S_{n+2} \mid R(w))$, knowing that $P(\varnothing, S_1 \mid R(w)) = 1$, where $\varnothing$ is the empty sequence. (Because $S_{n+2}$ is the unique final state of R(w) and, since $S_1$ is the unique initial state, we necessary begin in $S_1$ having produced an empty sequence.) Note that $P(v_1,\ldots,v_m, S_{n+2} \mid R(w))=P(v_1,\ldots,v_m \mid R(w))$. Since $P(v_1,\ldots,v_k, S_i \mid R(w))$ is the sum of the probabilities of each individual state sequence from the initial state to $S_i$ producing $v_1,\ldots,v_k$, we have the following recursion:

For all $1 \leq i \leq n+1$,

$$P(v_1,\ldots,v_k, S_i \mid R(w)) = P(v_1,\ldots,v_{k-1}, S_{i-1} \mid R(w))\, P(prod|phon(S_{i-1}))\, P(v_k|phon(S_{i-1}))$$
$$+ P(v_1,\ldots,v_k, S_{i-1} \mid R(w))\, P(del|phon(S_{i-1}))$$
$$+ P(v_1,\ldots,v_{k-1}, S_i \mid R(w))\, P(ins)\, P(v_k \mid ins) \qquad (2)$$

This formula expresses the fact that R(w) is in the state $S_i$, $i \leq n+1$, having produced the sequence of feature vectors $v_1,\ldots,v_k$ either:

- if R(w) was in the state $S_{i-1}$ having produced $v_1,\ldots,v_{k-1}$ and made a transition from $S_{i-1}$ to $S_i$ producing $v_k$ (phoneme production, for example, a French speaker sometimes produces the English phoneme /ð/ (in "**th**is") as the phone [z], or can pronounce it correctly by the phone [ð]); or
- R(w) was in the state $S_{i-1}$ having produced $v_1,\ldots,v_k$ and made a transition from $S_{i-1}$ to $S_i$ producing nothing (phoneme deletion, for example the /d/ in "an**d**" is often removed by many speakers); or
- R(w) was in the state $S_i$ having produced $v_1,\ldots,v_{k-1}$ and made a transition from $S_i$ to $S_i$ producing $v_k$ (phone insertion, for example an Italian often add a superfluous [ə] at the end of a word).

The final state has only one input transition, thus:

$$P(v_1,\ldots,v_k, S_{n+2} \mid R(w)) = (1-P(ins))\, P(v_1,\ldots,v_k, S_{n+1} \mid R(w)) \qquad (3)$$

where $P(v_1,\ldots,v_k, S_{n+1} \mid R(w))$ is computed by (2).

The computation of (2) is performed using the forward algorithm (See Problem 1 in Rabiner 1989). By starting with $P(\emptyset, S_1 \mid R(w)) = 1$, it is indeed possible to compute recursively the formula until producing $P(v_1,\ldots,v_m, S_{n+2} \mid R(w))$. The labels of the nodes in the lattice Figure 6 illustrate this process. $P(v_1,\ldots,v_m, S_{n+2} \mid R(w))$ then corresponds to the probability that R(w) produced the sequence of feature vectors $v_1,\ldots,v_m$.

Related to (1), recognizing a word from $v_1,\ldots,v_m$ therefore corresponds to computing:

$$\mathrm{Argmax}_w(P(v_1,\ldots,v_m, S_{n+2} \mid R(w)) \mid w \in \mathcal{W})$$

The problem is now to define all the free parameters of the model.

To establish a baseline against which to assess the efficacy of using a small set of words to establish a "sense" of how a particular speaker's accent restructures the native speaker's phonemes, we first model the process of **"naïve" recognition**, in which N assumes that when F pronounces a phone, it will lie close in feature space to N's phone.

For this aim, we need to define the initial state of every free parameter of the model:

- P(ins): the unconditional probability of inserting a feature vector,
- P(v|ins): the distribution over $\mathcal{V}$ of inserted feature vectors,
- P(del|p): the probability of deleting a given phoneme p,
- P(prod|p): the probability of producing a given phoneme p,
- P(v| p): the distribution over $\mathcal{V}$ for a given phoneme p to produce, corresponding to the map $\Psi_N$.

During naïve recognition, that is before any learning of a foreign accent, we use the following parameters.

P(ins)=0.01, corresponding to a low probability of insertion.

During an insertion, we consider that every feature vector can be inserted with the same probability, so P(v|ins) is taken to be uniform.

Remember that the outgoing transition probabilities from any state except the final one must sum up to 1. For all p in $\mathcal{P}$, we must therefore have P(ins)+ P(prod|p)+ P(del|p)=1. Since insertion, production and deletion are mutually exclusive, we have P(prod| p, ¬ins)+ P(del| p, ¬ins)=1 (if a transition is not an insertion, then it is either a production or a deletion). We chose to also set P(del| p, ¬ins)=0.01, thus leading to a low probability of deletion. We then have:

- P(del|p)=(1-P(ins)) P(del|p, ¬ins),
- P(prod|p)=(1-P(ins)) (1-P(del|p, ¬ins)),

ensuring that P(ins)+ P(prod|p)+ P(del|p)=1.

P(v| p), corresponding to the map $\Psi_N$, aims to define the probability over the feature space $\mathcal{V}$ for each native phoneme p in $\mathcal{P}$. For such a phoneme, we define the probability of perceiving the feature vector v = $(d_1,\ldots,d_k)$ as:

$$P\left(d_1,\ldots,\ d_k \middle| p\right) = \prod_{i=1}^{k} P\left(d_i \middle| p\right) \quad (4)$$

where $P(d_i|p)$ is the probability of perceiving the feature $d_i$ for a given phoneme p (features are defined in the "Feature space" section). The definition of the $P(d_i|p)$ elementary distributions are the following:

- For i = 1, $P(d_1|p)=1$ if $D_1(p)=d_1$, 0 otherwise. The idea is that vowel and consonant cannot swap (MacNeilage, 1998)
- For all the other features ($i \geq 2$), we assume that N allows a little uncertainty around a given feature for a given phoneme. We thus define the conditional distributions $P(d_i| p)$ by a (discretized) bell-shaped curve, a specialization of the Gaussian curve for discrete and bounded variables:

$$P(d_i|p) = \frac{e^{-\frac{(d_i - D_i(p))^2}{2\sigma^2}}}{N} \qquad (5)$$

where $D_i(p)$ is the value of the $i^{th}$ feature of the phoneme p, σ a standard deviation, and N is a normalization constant.

We made some assumptions here for the sake of simplification:

- We consider a Gaussian-like uncertainty whereas features are defined on bounded discrete variables, sometimes with very few values. The idea here is only to model the fact that the most probable perception of the $i^{th}$ feature ($i>1$) of a given phoneme p is $d_i = D_i(p)$, and the more $d_i$ moves away from this value, the more the probability decreases, according to the value of σ. As 99.7% of the values taken by data driven by a normal distribution are in the interval [μ-3σ, μ+3σ], where μ and σ are respectively the mean and the standard deviation of such a distribution, we choose to set the value of σ to 2/3, thus we can use the approximation that all the values expected for a given phoneme p are in the interval $[D_i(p)-2 , D_i(p)+2]$ (As an example, Figure 7 plots the distribution $P(d_i|p)$ for $i>1$ and p=/s/).
- We do not take yet into account the fact that some feature vectors cannot be produced (see the grey zones in Figure 3), which implies that features are not strictly independent from each other, from an articulatory point of view. The fact is that, as a simplification in the present study, we use phonetic features to model the input of speech perception, and it seems realistic to model a Gaussian-like uncertainty on each feature (which can be due to an environmental or a neural noise, for example). Moreover, the $P(v|p)$ conditional distributions we are defining in Equation 4 are initial parameters for naïve recognition, which will be then refined for accented speech adaptation (P3), allowing then to capture all the dependencies between features of a given phoneme.

The initial distribution $P(d_1,\ldots,d_k|p)$ for each native phoneme p, corresponding to the map $\Psi_N$, is then computed using (4).

*Exploiting Knowledge of the Word Being Pronounced (P2)*

The "naïve" process of the previous section simply seeks a word for which N's pronunciation approximates the sequence of feature vectors produced by F. However, the pronunciation of the agent F can be very different from what N would usually produce, and this can lead to a low word recognition rate with the naïve process. We thus turn now to a simple foreign accent learning process which allows the agent N to raise its recognition rate by approximating the map $H_{F\rightarrow N}$: $\mathcal{V} \rightarrow \mathcal{P}^*$ which is such that, for most words in $\mathcal{W}$, if $\Psi_F(\Phi_F(w))$ equals the string of feature vectors $v_1v_2 \ldots v_k$ used by F to pronounce w, then $\Phi_N(w)$ equals (more or less) the string of phonemes $H_{F\rightarrow N}(v_1)\ H_{F\rightarrow N}(v_2) \ldots H_{F\rightarrow N}(v_k)$.

What seems to happen in a real situation where a native speaker N is hearing a foreign one F is that N uses some easily recognizable part of the speech in order to infer a better model of F's pronunciation – N is able to learn to follow much of F's pronunciation from a relatively small training set.

In our HMM modeling framework, the approximation of the map $H_{F\rightarrow N}$ is done by tuning the free parameters described above, based on a training set $B_L$ comprising a set of (word, F's word pronunciation) couples, i.e., (w, $\Psi_F$ ($\Phi_F$(w))) pairs. Such a pair provides both the model (HMM) of the word w for N, R(w), and a sequence of feature vectors $v_1,\ldots,v_m$ from F's speech. These can be used to compute the most probable sequence of operations (substitution, insertion or deletion) in R(w) which produces $v_1,\ldots,v_m$. This is done using the Viterbi algorithm (Viterbi 1967), which is performed using virtually the same expression as in (2), just replacing the sum by a maximum. Indeed, we want to find the path in the derived lattice of R(w) (see example Figure 6) which maximizes the product of the weights of the arrows traversed on this path. For this aim, let us define the quantity:

For all $1 \leq i \leq n+1$:

$$Q(v_1,\ldots,v_k, S_i \mid R(w))$$

$$= \max( Q(v_1,\ldots,v_{k-1}, S_{i-1} \mid R(w))\, P(prod|phon(S_{i-1}))\, P(v_k|phon(S_{i-1})),$$

$$Q(v_1,\ldots,v_k, S_{i-1} \mid R(w))\, P(del|phon(S_{i-1})),$$

$$Q(v_1,\ldots,v_{k-1}, S_i \mid R(w))\, P(ins)\, P(v_k \,|ins) ) \qquad (6)$$

This formula is computed using the forward algorithm as for (2), beginning with $Q(\emptyset, S_1 \mid R(w)) = 1$. But at each step of the recursion $Q(v_1,\ldots,v_k, S_i \mid R(w))$, what we call the step (k,i), we save a pointer to the step corresponding to the maximum in (6), that is:

- (k-1, i-1) if the maximum is the first term in (6),
- (k, i-1) if the maximum is the second term in (6),
- (k-1, i) if the maximum is the third term in (6),

until the step (m,n+1). As one and only one pointer is associated which each step, we can trace back the pointers from (m,n+1) to (0,1). This gives the most probable path into the lattice to produce $v_1,\ldots,v_m$, hence the most probable sequence of operations in R(w) which produces $v_1,\ldots,v_m$.

### *Refining the free parameters of the model (P3)*

This path thus provides a set of state transitions and feature vector emissions for each word of the training set, allowing us to compute:

- N(ins), the number of insertion transitions, and N(¬ins), the number of non-insertion transitions (i.e. production or deletion transitions) in the training set,
- N(del|p), the number of deletion transitions for a given phoneme p, and N(prod|p), the number of production transitions for that phoneme,
- $N(v_i|ins)$, the number of occurrences of each inserted feature vector $v_i$ during insertion transitions,
- $N(v_i|p)$, the number of occurrence of each produced feature vector $v_i$ during production transitions for a given phoneme p.

To update the free parameters of the model, we then use a classical generalization of the Laplace succession rule using a non-uniform prior (Jaynes 2003, chapter 18), which allows us to combine both the initial parameters (those described for naïve recognition) and the information from the training set. The updated parameters are defined hereunder, where C is the weight of the initial parameters relative to the data from the training set:

- $P_{upd}(ins) = \dfrac{N(ins) + C.P(ins)}{N(ins) + N(\neg ins) + C}$

- $P_{upd}(del \mid p, \neg ins) = \dfrac{N(del \mid p) + C.P(del \mid p, \neg ins)}{N(del \mid p) + N(prod \mid p) + C}$, used to compute:

  - $P_{upd}(del|p)=(1-P_{upd}(ins))\ P_{upd}(del|\ p, \neg ins)$
  - $P_{upd}(prod|p)=(1-P_{upd}(ins))\ (1-P_{upd}(del|\ p\ , \neg ins))$

- $P_{upd}(v_i \mid ins) = \dfrac{N(v_i \mid ins) + C.P(v_i \mid ins)}{C + \sum_{v_j \in V} N(v_j \mid ins)}$

- $P_{upd}(v_i \mid p) = \dfrac{N(v_i \mid p) + C.P(v_i \mid p)}{C + \sum_{v_j \in V} N(v_j \mid p)}$

In the result section, we will choose C=20.

## Results

We introduced $H_{F\rightarrow N}$: $\mathcal{V} \rightarrow P^*$ as the probabilistic map such that, for most words in $\mathcal{W}$, if $\Psi_F(\Phi_F(w))$ equals the string of feature vectors $v_1v_2 \ldots v_k$ used by F to pronounce w, then $\Phi_N(w)$ equals (more or less) the string of phonemes $H_{F\rightarrow N}(v_1)$ $H_{F\rightarrow N}(v_2) \ldots H_{F\rightarrow N}(v_k)$. We now assess how much the learning of the map $H_{F\rightarrow N}$ during a learning phase is able to improve the recognition rate during a test phase. We assess the reliability of the $G_{F\rightarrow N}$ function that we defined in the Principles section – for each string θ of feature vectors that is actually pronounced by F, $G_{F\rightarrow N}(\theta)$ is the word of English that N understands F to have intended to say by uttering this string. (The map is probabilistic.)

In the elicitation paragraph of our corpus (69 words), we define the training set as the first 35 words and the test set as the 34 remaining words. We run the word recognition process on the test set a first time without foreign accent learning (parameters defined for “naïve recognition”), and a second one after running the foreign accent learning process on the training set. The question is then “how much is the learning process on the training set able to improve the recognition rate on the test set?” We first analyze how the model is able to extract the information contained in the training set, then show how this new information improves the word recognition in the test set.

### *Training phase*

Here we present the effect of the learning phase on a particular French speaker of English, French8 (Weinberger 2010). In the “Corpus” section, we earlier presented the transcription of the elicitation paragraph as spoken by the native agent N whose pronunciation is given by the MRC Psycholinguistic Database, as well as the phonetic transcription for French8 (F).

Table 3 compares the set of productions, insertions and deletions which allows us to transform N’s phonetic transcription into F’s one, both “manually” (when a human look up the transcriptions, column 3 and 4) and as performed by the model using the Viterbi algorithm as defined by equation (6).

Table 3 shows that:

- Most transformations involved in the learning base are also involved in the test set,
- The model, using the Viterbi algorithm as defined equation (6), is able to detect most transformations of the learning set.

The only failure of the model is that it is unable to detect the mispronunciation of the native English phoneme /ɹ/ (pronounced either as the phones [r] or [ʀ]). This comes from the fact that the native phoneme and the pronounced phone are too far one from another in the feature space (see Figure 3). As a result the model finds that a deletion and an insertion is more likely than a substitution.

### *Testing phase*

Further information provided by Table 3 is that many of the transformations found in the training set are also involved in the test set. Therefore we can expect that the learning phase can improve the recognition rate during the test phase. It is indeed the case when we compare the recognized words in the test set before and after learning.

Here are the words recognized by the model for French8 transcription, before accent learning (naive recognition):

[ALSO - ] --- [NEED - ] --- [OR - ] --- [SMALL - ] --- [PLASTIC - ] --- [SNAKE - ] --- [AND - ] --- [OR - ] --- [PICK - ] --- [TOY - ] --- [WALK - ] --- [VAGUE - ] --- [DOOR - ] --- [KIDS - ] --- [SHE - ] --- [CAN - ] --- [SCOOP - ] --- [NICE - ] --- [TAKES - ] --- [INTO - ] --- [TEA - T - ] --- [GET - ] --- [BAGS - ] --- [AND -

] --- [WE - ] --- [WILL - ] --- [GO - ] --- [MEET - ] --- [HANG - ] --- [WEDNESDAY - ] --- [AT - ] --- [TO - ] --- [TAKEN - ] --- [STATION - ] ---

Two words in the same brackets means that they are recognized with the same probability.

Many of these words are misrecognized (recognition rate around 62% of the test set).

After foreign accent learning on the training set, the recognized words are the following:

[ALSO - ] --- [NEED - ] --- [A - ] --- [SMALL - ] --- [PLASTIC - ] --- [SNAKE - ] --- [AND - ] --- [A - ] --- [BIG - ] --- [TOY - ] --- [FROG - ] --- [FOR - FOUR - ] --- [THE - ] --- [KIDS - ] --- [SHE - ] --- [CAN - ] --- [SCOOP - ] --- [THESE - ] --- [THINGS - ] --- [INTO - ] --- [TEA - T - ] --- [RED - READ - ] --- [BAGS - ] --- [AND - ] --- [WE - ] --- [WILL - ] --- [GO - ] --- [MEET - ] --- [HER - ] --- [WEDNESDAY - ] --- [AT - ] --- [THE - ] --- [TRAIN - ] --- [STATION - ] ---

Almost every word is now correctly recognized (97% of the test set). Therefore, the information contained in training set about insertion, substitution and deletion seems to be accurate in the test set.

We ran this process (word recognition on the test set, before and after learning) on two different groups of transcriptions provided by the corpus. Group A contains 10 native English speakers (English onset at birth, naturalistic English learning method, length of English-speaking residence greater than 15 years). Group B contains 10 foreign speakers of English (English onset greater than 15 years, academic learning method, English residence less than 1 year). Figure 8 shows the results. The details of recognized words before and after learning for each speaker of groups A and B are provided as a supplementary material.

Regarding group A of native speakers, we observed a recognition rate around 80% of the test set words before and after learning. For group B of foreign speakers, we observed a worse recognition rate before learning than for group A, but the same rate after learning.

We performed a two-way ANOVA to analyse the effect, on the recognition rate, of learning (with two conditions: before vs. after learning) and speakers (with two conditions: native vs. foreign speakers) as well as their interaction. We found a highly significant effect:

- of learning ($F(1,36)=29.2$, $p<0.0001$), with a recognition rate significantly larger after learning than before;
- of speakers ($F(1,36)=7.3$, $p<0.02$), native speakers being better understood than foreigners;
- and of their interaction ($F(1,36)=8.13$, $p<0.01$), the effect of learning being more important on foreign speakers than on native ones.

The simple foreign accent learning process we propose therefore provides insight into the key observation in the introduction: when we meet someone with a novel accent, we may at first understand relatively little of what they say, yet after attending closely to a few sentences, we may thereafter follow their speech with greatly improved accuracy.

However, the model we have offered is completely neutral as to whether the features which characterize the phones produced by N and by F are motor or perceptual or a mixture thereof. In particular, then, the model as it stands has no direct relation to either mirror neurons or to the motor theory of speech perception.

Part 3, then, will reconcile the computational processes involved in the model with those of the neurolinguistics analysis of Part1, extracting some implications for the reframing of the motor theory.

## Part 3. Reframing the Motor Theory

Part 1 offered a preliminary reconsideration of the motor theory and the role of mirror neurons in speech perception that is informed by the perceptuo-motor view of the PACT and earlier models of the adaptive formation of mirror neurons for grasping. It allows viewing extensions of that mirror system as part of a larger system for neuro-linguistic processing.

Then in Part 2, we implemented an HMM-based computational model, analyzing the general challenges of solving the problem of adapting recognition to foreign-accented speech, thus providing new insight into the key observation in the introduction: When we meet someone with a novel accent, we may at first understand relatively little of what they say, yet after attending closely to a few sentences, we may thereafter follow their speech with greatly improved accuracy.

We now offer a novel synthesis of Figure 1, which offered an account of mirror neurons within an evolution-grounded overview of the dorsal and ventral paths for language perception and production, and Figure 5, which relates our HMM-based computational model to the three problems a HMM is basically able to solve. We use this integration (Figure 9) to reframe the motor theory.

## A Unifying Framework for Modeling

How, then, might this HMM-based model be interpreted in relation to the ventral and dorsal streams of Part 1 (Figure 1), and what would then correspond to mirror neuron training? A key point in what follows is to bear in mind the distinction between two tasks:

- Imitating what the speaker says in the speaker's accent: This requires recognizing the motor gestures *of the speaker*.
- Recognizing what the speaker is saying and/or repeating the words in one's own accent: This does not imply recognizing the motor gestures *of the speaker,* but may involve access to the own motor programs of the listener.

The upper part of Figure 9 provides a simple view of the dorsal pathway. At the top, auditory analysis may activate mirror circuits at the level of phones which can then activate articulators to produce them, and this serves as the basis for imitating non-words and novel words. (Not shown in the diagram, there are also processes for extending the stock of native phones to include approximations to those used in non-native words or native words in a foreign accent.) Below this, we show that in some cases (especially with discourse context) auditory input to the dorsal path can activate mirror circuits for familiar words and these can in turn drive articulators for these words. Bidirectional arrows suggest cross-coupling between the two dorsal paths but we do not develop this notion further in this paper. The articulation of words may become overt (as in imitation), or may be inhibited while still generating corollary discharge for phones (P) and words (W).
We have diagrammed the various HMM processes studied in Part 2 as occurring in the ventral stream. The point to bear in mind is that here a word w is being extracted as an access point to semantics -- but we show a sidepath to speech production which passes to the articulators for the word shown in the dorsal pathway. "Semantics" can similarly direct articulation of words related to an encoded meaning. But now let's relate the ventral pathway to the three basic problems that an HMM is able to solve (Rabiner 1989) and the form in which we developed algorithmic implementations in Part 2. Noting that we use English as our exemplar language, without restricting the applicability of the general theory, we recall the notations $\mathcal{P}$ (the set of English phonemes), $\mathcal{V}$ (a feature space for phones), $\mathcal{W}$ (a set of words of English) and $\mathcal{P}^*$ (the set of strings of English phonemes).

For now, ignore module "A?". It is an alternative to "A" which we will discuss later.

***Case P1:***

Given an observation sequence in $\mathcal{V}^*$, what is the probability that it comes from a given HMM R(w)? In Figure 5, P1 corresponds to the model that can already recognize words when they are pronounced in accent N. In Figure 9, this is the path through A to Competing HMMs when the listener assumes that the input is in her native accent, thus interpreting the auditory input as a feature vector to be turned into weighted candidates for native phonemes. These are then supplied to each HMM which computes the probability that the input sequence could have been a (mis)pronunciation of its word. The winner of the competition is the recognized word.

The speaker effect revealed by the ANOVA in Part 2 (Figure 8) shows, within the simplifications of our model, that this process of "naïve recognition" is less efficient for foreign-accented speech than for the listener's "native" pronunciation.

***Case P2***

P2 maps $\mathcal{V}^*$ x $\mathcal{W}$ into $\mathcal{P}^*$. Given an observation sequence in $\mathcal{V}^*$ and an HMM R(w), what is the most likely state sequence in $\mathcal{P}^*$ the observations come from? In Figure 5, the model is able to use a few easily recognizable words to segment a new accented speech from F into a phoneme sequence. P2 corresponds to using the top-down bias from ongoing discourse context to suggest that a given word was the basis for the auditory input, thus providing data for matching the input feature vectors against native phoneme candidates.

The learning effect revealed by the ANOVA shows that using lexical hypotheses on a small word set (35 words in our case) provides enough information about F's pronunciation to generalize to novel words and thus significantly improve the recognition rate.

***Case P3:***

Given an observation sequence in $\mathcal{V}^*$ and an HMM R(w), how can the free parameters of the model be refined? In Figure 5, this segmentation and the knowledge of the auditory input underlying it allows to refine the parameters of the model (training feedback P3), gaining fast access to a reasonably accurate level of recognition in the new accent F.

The output choice from Competing HMMs serves to train the accent system A (which represents the current state of P2 for each known accent). It provides the training feedback for P3 to update the way in which, dependent on the current speaker/accent code which is provided by a separate system that recognizes the speaker or accent, System A transforms auditory input to a probability weighted distribution on native phonemes. As training proceeds, the system is able to recognize words in any particular foreign accent on which A has been trained with greater accuracy.

The interaction effect between learning and speakers revealed by the ANOVA shows that this learning process is especially useful for foreign-accented speech. Moving beyond the model of Part 2, we add pathway X (which may be modulated by discourse context) to record the fact that in some contexts some words may readily be recognized on the basis of part of the auditory input without segmentation down to the level of phone candidates.

***Dorsal assistance to the ventral pathway***

Let us turn now to the availability to the ventral path of corollary discharge from the motor pathway for phones (P) and words (W). The corollary discharge P is made available to the "accent box" A, while W is made available to the competing HMMs. If the top contenders among the competing phone candidates or HMMs are either very close or all very low in estimated probability, then the winner-takes-all process among the competing HMMs will be accordingly slow down to gather more input to decide on which word is to be recognized. This leaves time for supplementary processes to take place. The corollary discharges P and W for the leading contenders can then drive comparison with the auditory feedback which can then tip the balance. As noted, this process is especially important when the input is noisy or otherwise degraded.

Finally, we note two other features of Figure 9. One is that both the recognition of a word or the semantics of a word can drive speech production via the dorsal path (recall Figure 1). Secondly, it may be that the processes described in A are not located within the ventral pathway per se, but instead are located (as suggested by the box A?) in such a way as to provide preprocessed input for both the dorsal and ventral pathways. Such an A? would have the speaker/accent code set to "native" when the task is to imitate, with more or less accuracy, the actual sounds produced by the foreign speaker, whereas it would be set to the code for that speaker/accent if the task is to articulate the word that may have been intended by the foreign speaker, but in the native's accent. Note that the latter task can be accomplished both via the dorsal path alone (when the focus is on decoding the sound of the input word) and via the ventral path when the focus is on the word as a meaningful unit rather than as an articulatory template.

## Assessing Claims for the Motor Theory and Mirror Neurons in Speech Perception

With this background, we can assess the extent to which the motor theory of speech perception is correct.

With Figures 1, 5 and 9, we have completed our task of reframing the motor theory of speech perception in a way which integrates it with certain key processes of speech production. As a result, we see there are important tasks (learning new words, imitating foreign accents, compensating for noise and distortion) in which motor representations play an important role, as documented in papers in which the motor theory was revised (Liberman and Mattingly 1985) and reviewed (Galantucci et al. 2006). However, at the same time we have shown how to accommodate observations (Lotto et al. 2009) showing cases in which motor representations do not play a crucial role.

Intriguingly, our model variant in which A is moved to A? suggests that motor representations may play an important role even in our paradigm case of processing speech in a foreign accent, but now the motor gestures related to speech perception are not (save in the case of imitation of how the word is pronounced in a foreign accent) those of the speaker, but rather those the listener would use to pronounce the word in her accent.

Chinchillas can discriminate some aspects of human speech (Kuhl and Miller 1975) so that a production system is not necessary for recognition of these aspects to occur. In the cited study, four chinchillas were trained to respond differently to syllables with /t/ and /d/ consonants produced by four speakers in three vowel contexts. But discriminating a couple of phonemes is not quite the same as disentangling a fast verbal performance into a hierarchical structure of words to extract the meaning of the sentence. Turning to coarticulation, when synthesized stop consonants varying perceptually from /da/ to /ga/ are preceded by /al/, human listeners report hearing more /ga/ syllables than when the members of the series are preceded by /ar/. Is this the result of specialized processes for speech that compensate for acoustic consequences of coarticulation? Lotto et al. (1997) trained Japanese quail to peck a key differentially to identify clear /da/ and /ga/ exemplars. After training, ambiguous members of a /da/-/ga/ series were presented in the context of /al/ and /ar/ syllables. Pecking performance demonstrated a shift which coincided with data from humans. Yet quail cannot produce the actions for human speech. Lotto et al. thus suggest that processes underlying "perceptual compensation for coarticulation" are species-general, not the result of a human-specific “speech module”. But it is unclear to us why a bias of /da/-/ga/ discrimination should be viewed as "perceptual compensation for coarticulation" rather than the converse. Moreover, it is possible to argue for speech-specific circuitry while agreeing that it cannot be as encapsulated in a module as envisioned in the revised version of MT (Liberman and Mattingly 1985). The lesson of Figure 1 is that the same sensory data (auditory, in the present example) can be processed in different systems. Thus the binary classification of elements of a /da/-/ga/ series in varied contexts by quail suggests that it corresponds to a general auditory decision making system whose highly processed inputs (recall the role of cIPS in the FARS model) would then also become available to mirror neurons as the motor control system for speech evolved along the human line, serving to constrain what patterns of coarticulation would be favored by the speech apparatus.

It is clear that for humans to get speech, the vocal apparatus and its control had to evolve away from that of our common ancestors with other nonhuman primates. It also appears that a concomitant expansion of auditory working memory was required, something that also seems to characterize those male songbirds that learn a complex song before being able to produce it. However, where many songbirds have only a limited repertoire of songs with no linkage of song structure to novel semantic structures, humans are constantly able to recognize and produce novel and quite lengthy utterances. This fits in with the stage of complex action recognition and imitation which is part of the elaboration of the Mirror System Hypothesis. In pursuing its relevance to spoken language, and in the spirit of Figure 1, we reiterate that the child learning to speak (and the adult learning a new word) must attend to the "metrics" of the articulatory action and seek to reproduce it, perhaps crudely at first but with increasing fidelity – perhaps using an auditory template akin to that used by the young songbird. In general, though, the human's learning, unlike that of a songbird, is not of a sound pattern alone (though it might be) but at the same time involves acquiring a meaning for the word – thus adapting synapses both within the "mirror for words" and the ventral pathway in Figure 1.

The sound that an audiovisual mirror neuron learns to respond to has no a priori relation to the visual or motoric structure of the action. The learning theory of mirror neurons implies that mirror neurons do not start with the mirror property but as quasi-(or pre-) mirror neurons which get started by responding to corollary discharge of a self-action and end by responding also to sensory input related to execution of the action. Imitation requires different mechanisms not in the MNS model that, among primates, are most developed in humans– responding to sensory input and as a result shaping motor activity whose corollary discharge yields similar sensory input.

This is to be contrasted with sensorimotor associations in which one learns a map from sensory inputs to motor outputs, but with no requirement that the motor act produces sensory inputs akin to the stimuli – as in pushing one lever in response to a red stimulus, another in response to a blue one – the mechanism that, presumably, underlies the quail data of Lotto et al. (1997). The fact that a sensorimotor association system exists in quails need not preclude the existence of a mirror system for speech sound in humans. This is neutral as to whether or not there is a separate neural module. Just as context can switch the meaning gleaned from a given sound like that of /row/, so can the expectation that something is a speech sound change its interpretation – but whether this involves priming of distinct circuitry or modulation of neurons within a given circuit remains an open issue. (See the discussion in Arbib 2006; addressing cases of lesions which dissociate apraxia and aphasia as cited by Barrett et al. 2005 -- suggesting how a basic set of circuitry may be conserved while circuits with new functionality evolve "atop" it.)

Consistent with this, PACT does not consider motor simulation as the sole and key process underlying speech recognition. Basically, the assumption in PACT is that perceptual processing occurs in the ventral route (organized through speech learning in a way that does interact with motor representations in the dorsal route), and that the dorsal route can provide access to motor representations for organizing the speech scene, extracting relevant information, segmenting speech, complementing information that is masked or lacking, and processing of speech in a foreign accent, etc. Data from the Grenoble lab (Basirat et al. 2008; Sato et al. 2004) on the role of the parieto-frontal pole for segmentation in verbal transformations (e.g. from "…life life life …" to "… fly .fly fly …") is of interest here.

Damage to left frontal and posterior parietal regions (with sparing of Broca's area, superior temporal gyrus and the tissue in between) seems to disrupt networks playing a part in mapping speech onto conceptual-semantic representations while leaving the sensory-motor functions that support repetition of speech intact. Lotto et al. (2009) comment that this dissociation is opposite to the deficits of Broca's aphasia, indicating that – "directly counter to MT" –preservation of motor speech functions is neither necessary nor sufficient for speech perception. However, we repeat the claim that two different processes are at work – a ventral path: phonological access to word form which can access the semantics of the word; and a dorsal path which "directly" accesses motor forms for phones (or larger speech units such as syllables or moras which incorporate coarticulation). The phonological word form can be linked to a dorsal process for articulating the word either directly or via the semantic form (which may well be associated with multiple words and phrases with similar meanings). The revised version of MT (Liberman and Mattingly 1985) claims that (i) processes of speech motor planning are mandatory to speech perception, and (ii) the shared representations of speech perception and production are articulatory (motor) and linguistic. As Lotto et al.

(2009) show, there is little evidence for a mandatory role for production processes in speech perception – but we stress that impairment of motor output can impair speech perception (e.g., Luria 1973). As they concede, it is thus possible that production can aid perception, especially in challenging listening situations. When doubts need to be resolved, production processes could be used to create representations of candidate words or syllables to be compared to the auditory input, as in analysis by synthesis models (Skipper et al. 2007; van Wassenhove et al. 2005), and this may rest on a mirror system loop involving inverse/forward model pairs (as first proposed by Arbib and Rizzolatti 1997). However, this interaction need not be required for normal "low load" speech perception. For example, Moineau et al. (2005) investigated individuals with Broca's aphasia. They used a combination of acoustic distortions to probe lexical comprehension and concluded that accurate and efficient lexical processing that processing under these conditions reveals the continuous nature of the impairment of linguistic behaviors observed in individuals with aphasia. Thus mirror neurons in Broca's area cannot be the "be-all and end-all" of word recognition but nonetheless may contribute within the framework of "competition and cooperation" of multiple pathways.

# FIGURES

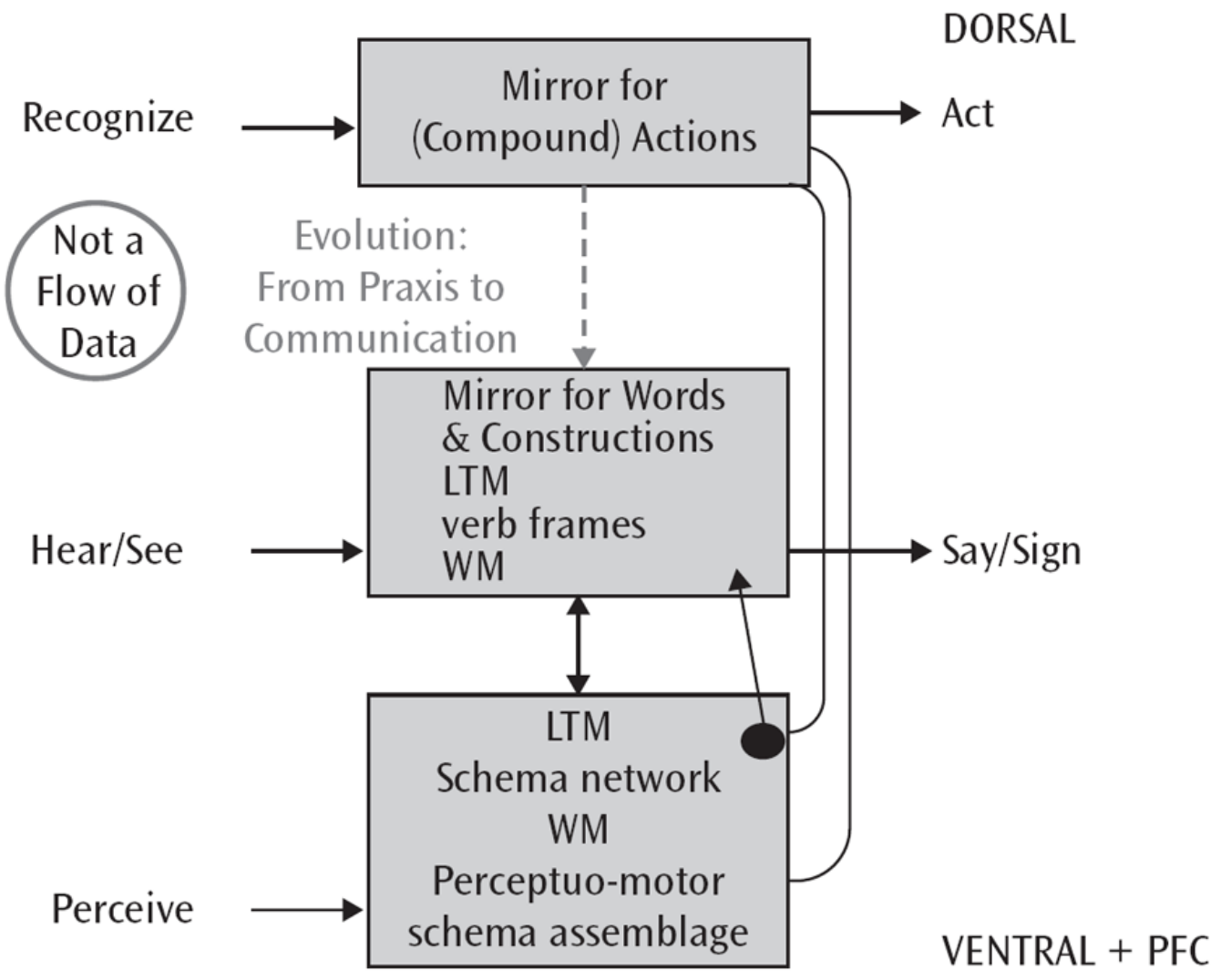

Figure 1: A view of neurolinguistic process informed by the Mirror System Hypothesis. Words as signifiers (articulatory actions) link to signifieds (schemas for the corresponding concepts), not directly to the dorsal path for actions (based on Fig. 4 from Arbib, 2006). The ‘‘mirror systems” are posited to contain not only mirror neurons but also canonical and other neurons. Then, extending the basic scheme for single actions and words, we employ complex imitation to lift execution and observation from single familiar actions to novel compounds, and similarly lift words to more complex utterances via the use of constructions.

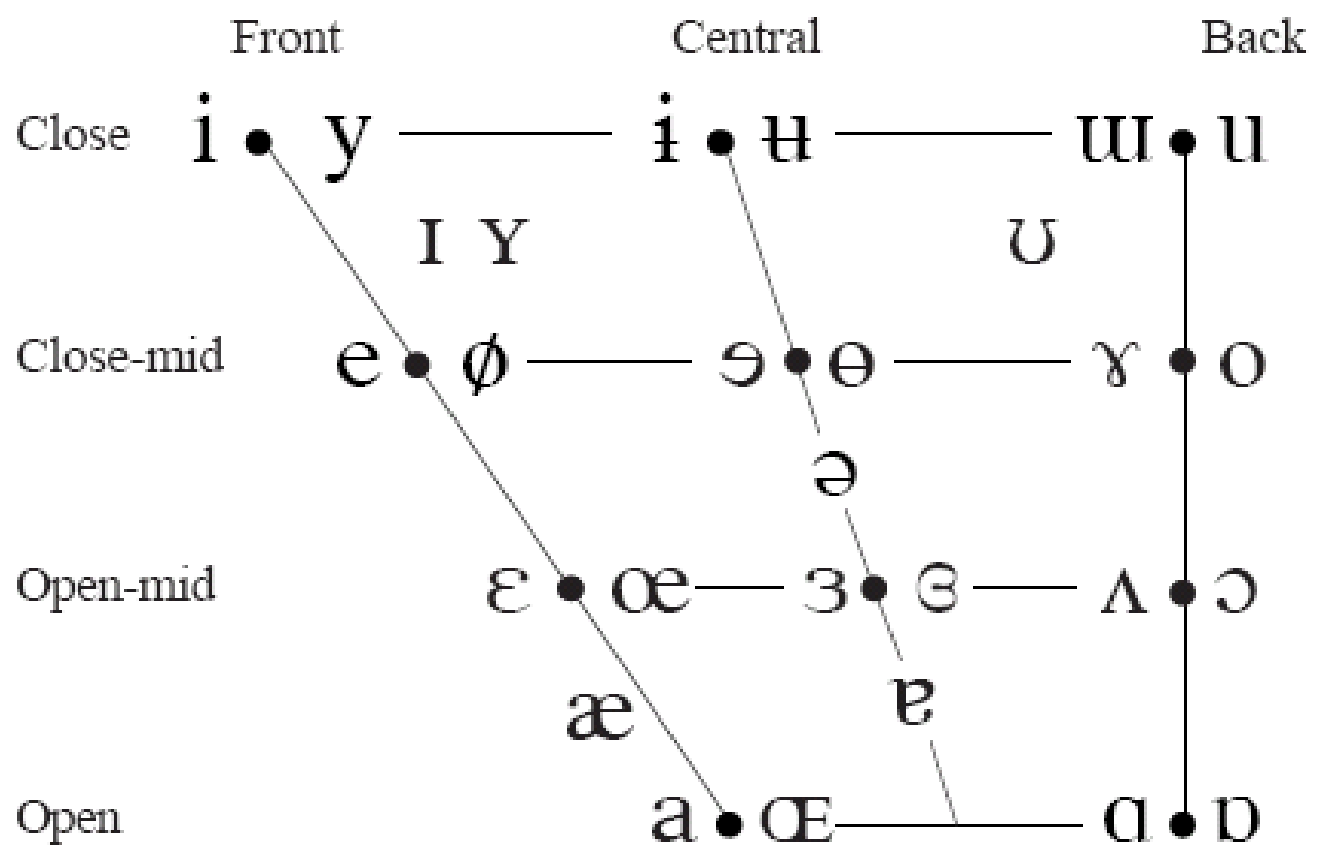

Figure 2: Main vowel features from the International Phonetic Association (1999). The figure makes clear the Front/Back and Open/Close dimensions. When symbols appear in pairs, the one to the right represents a rounded vowel. The figure does not show the nasalization dimension – each vowel shown here can be pronounced with or without nasalization.

| | Bilabial | Labiodental | Dental | Alveolar | Postalveolar | Retroflex | Palatal | Velar | Uvular | Pharyngeal | Glottal |
|---|---|---|---|---|---|---|---|---|---|---|---|
| Plosive | p b | | | t d | | ʈ ɖ | c ɟ | k g | q ɢ | | ʔ |
| Nasal | m | ɱ | | n | | ɳ | ɲ | ŋ | ɴ | | |
| Trill | ʙ | | | r | | | | | ʀ | | |
| Tap or Flap | | ⱱ | | ɾ | | ɽ | | | | | |
| Fricative | ɸ β | f v | θ ð | s z | ʃ ʒ | ʂ ʐ | ç ʝ | x ɣ | χ ʁ | ħ ʕ | h ɦ |
| Lateral fricative | | | | ɬ ɮ | | | | | | | |
| Approximant | | ʋ | | ɹ | | ɻ | j | ɰ | | | |
| Lateral approximant | | | | l | | ɭ | ʎ | ʟ | | | |

Figure 3: Main consonant features from the International Phonetic Association (1999). The horizontal axis is the Place dimension, ranging from glottal to bilabial, while the vertical axis is the Manner dimension, ranging from lateral approximant to plosive. Where symbols appear in pairs, the one to the right represents a voiced consonant. Shaded areas denote articulations that appear to be impossible. Considering the phoneme /w/, which is a voiced approximant with a double place of articulation (labial-velar), and is hence not represented in Figure 2, we decided to encode it as a voiced bilabial approximant.

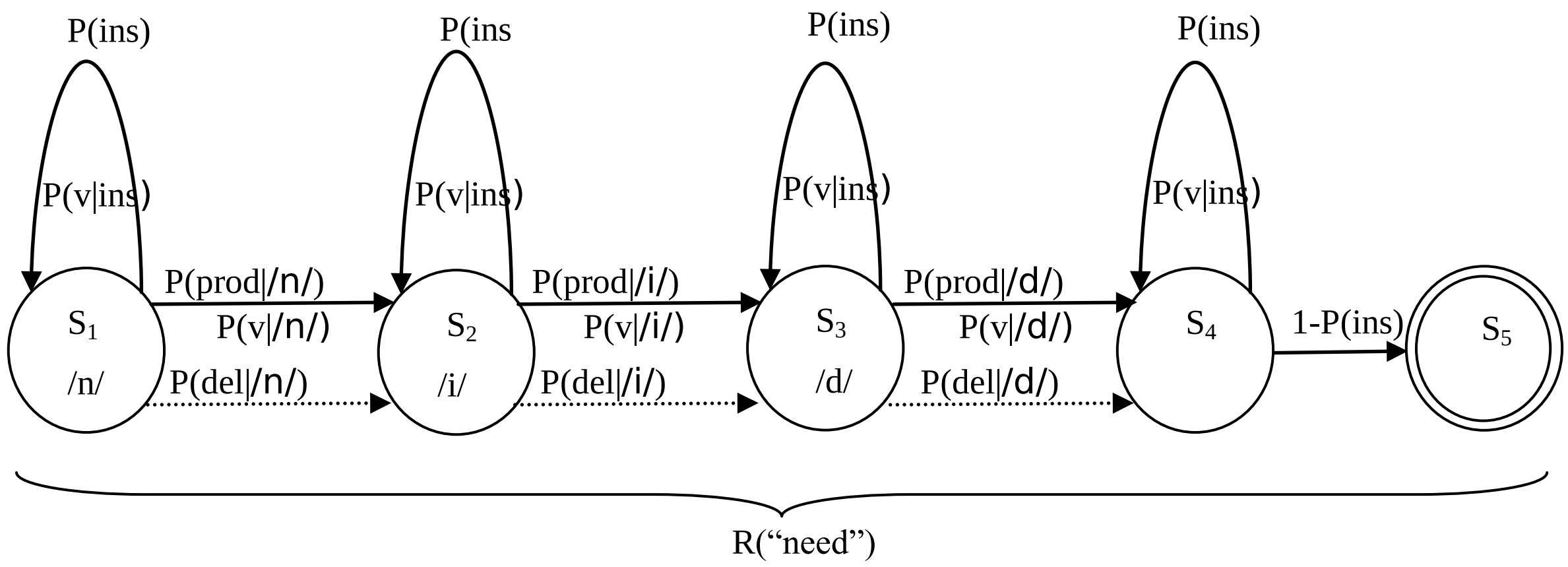

Figure 4: The HMM R("need") for the word "need" with $\Phi_N$(need) = /nid/. The unique initial state is $S_1$, the unique final state is $S_5$. For each state $S_i$, $1 \le i \le 3$, we write the corresponding phoneme phon($S_i$) of the phonetic transcription $\Phi_N$*("need")* = /nid/. The probability of each transition is given above the arrow representing it. The output probability of producing a feature vector during a transition is given below the arrow representing it (except in the case of deletion (dotted arrows) and in the transition to the final state $S_5$, where the HMM produce nothing with probability 1).

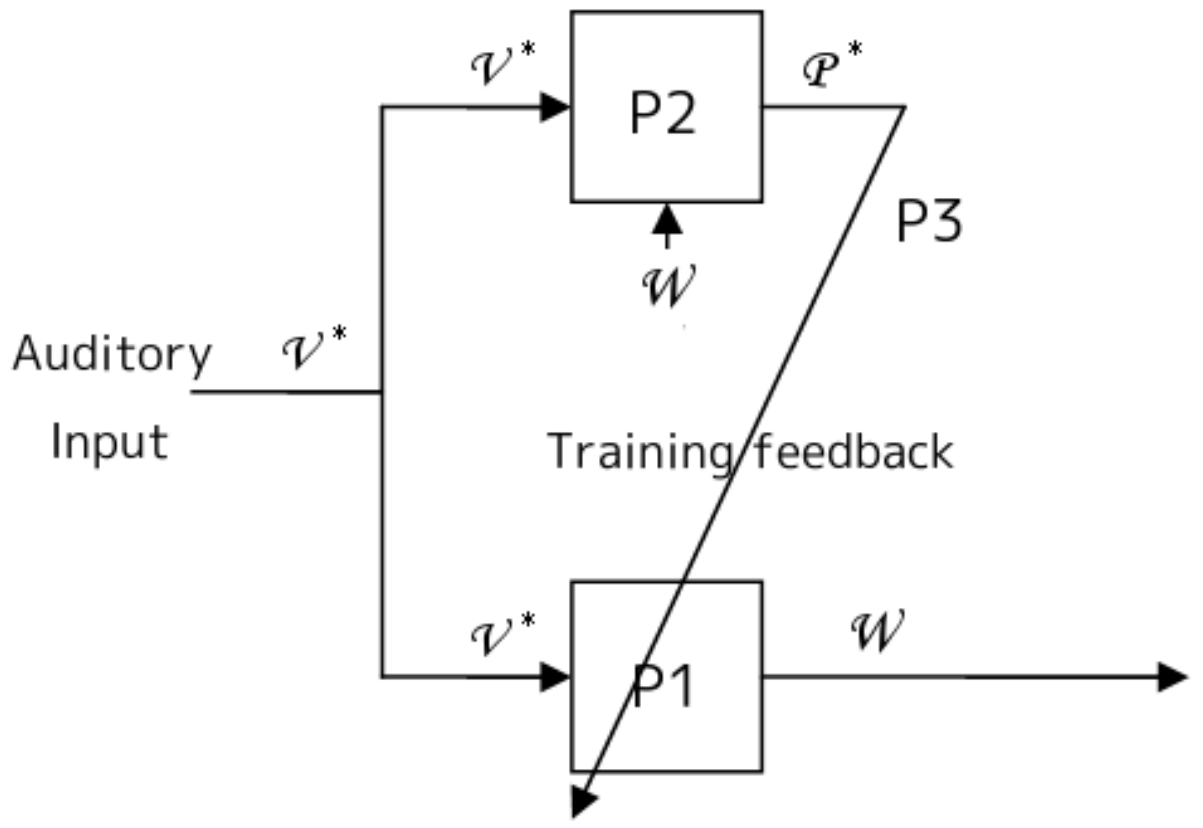

Figure 5: The three problems that an HMM is able to solve in terms of our accent model. Given that the model can already recognize words when they are pronounced in accent N (P1, from V * to W ), the model is able to use a few easily recognizable words to segment a new accented speech from F into a phoneme sequence (P2, from (V *x W ) to P*). This segmentation and the knowledge of the auditory input underlying it allows the model to refine the parameters of the model (training feedback P3), gaining fast access to a reasonably accurate level of recognition in the new accent F.

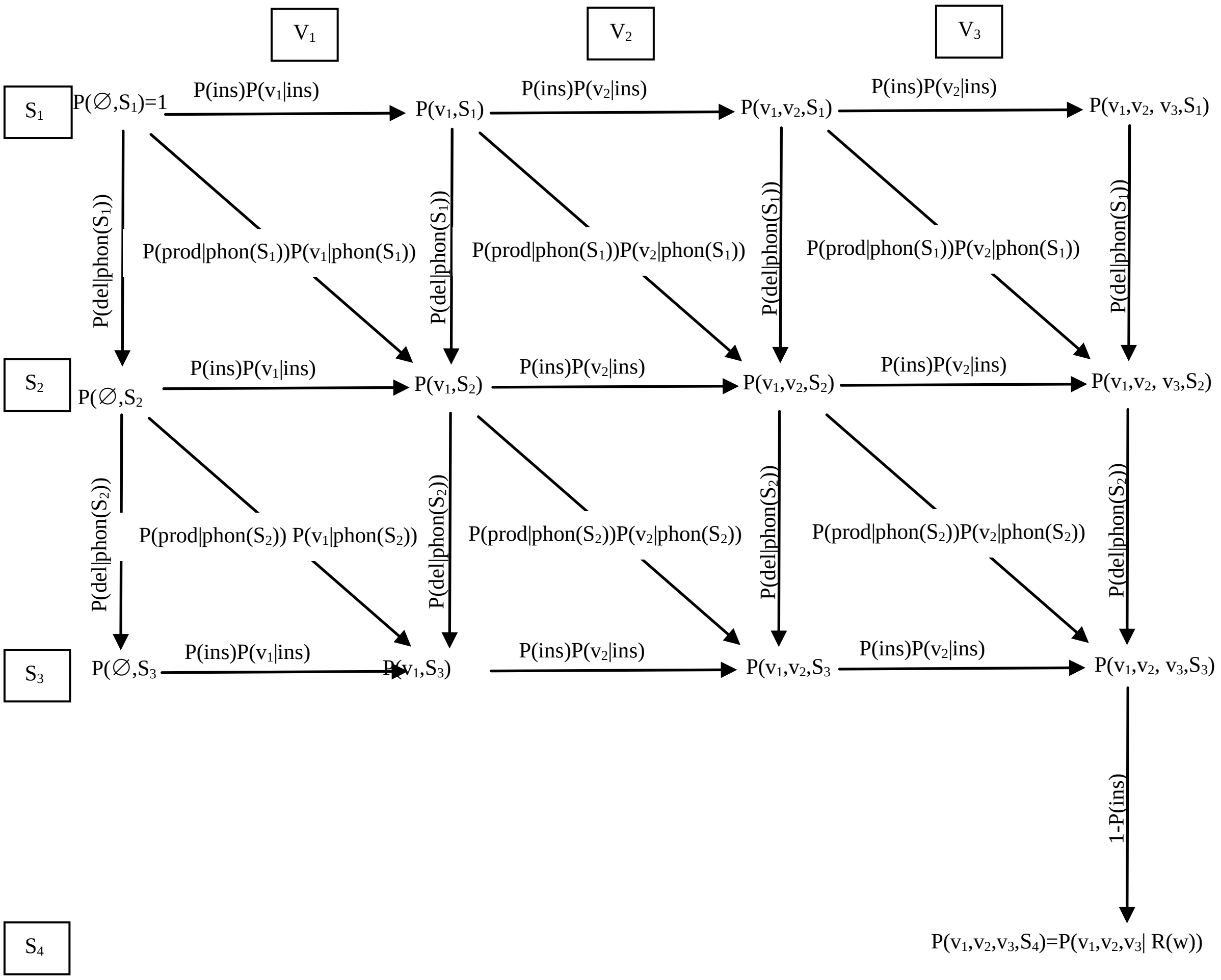

Figure 6: Graphical representation of each of the possible state sequences in R(w) which can produce $v_{1,}\ldots,v_m$, with w having 2 phonemes (the HMM R(w) therefore having 4 states, from $S_1$ to $S_{4,}$) and m= 3. Each arrow of the lattice corresponds to a transition (horizontal: insertion; vertical: deletion; oblique: production), weighted with the probability of doing it and emitting the corresponding feature vector (or emitting nothing in the case of vertical transitions). Each node is labeled by $P(v_1,\ldots,v_k, S_i)$, the probability of being in the state $S_i$ having produce $v_{1,}\ldots,v_k$ (∅ is the empty sequence).

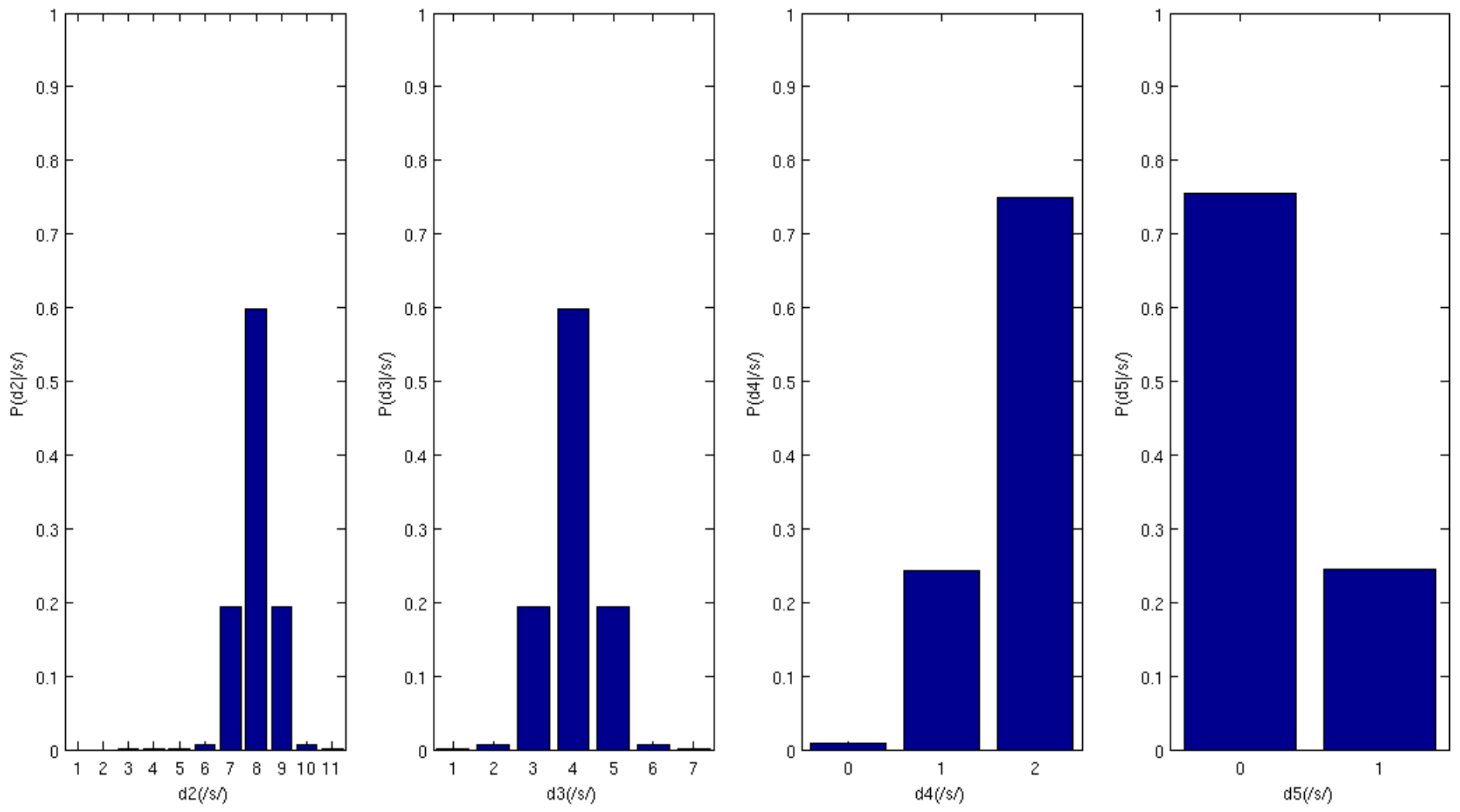

Figure 7: Initial distribution $P(d_i|p)$ for i>2 and p=/s/, computed by (5). $d_2$ is the place dimension, from glottal (1) to bilabial (11); $d_3$ is the manner dimension, from lateral approximant (1) to plosive (7); $d_4$ is the voiced dimension, from nasal (0) to non-voiced (2); $d_5$ is the nasal dimension, from non-nasal (0) to nasal (1).

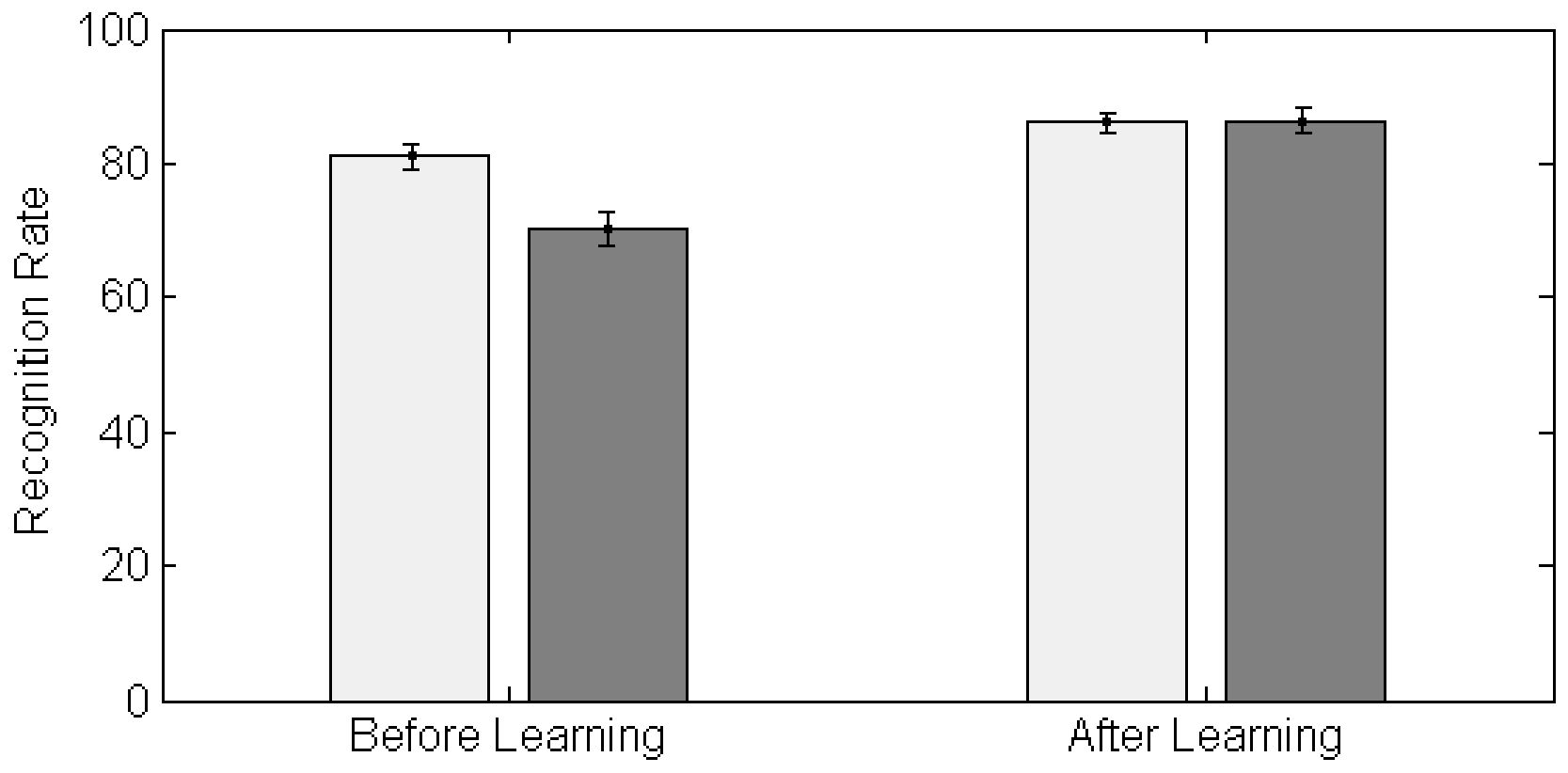

Figure 8: Average recognition rate on group A (light grey) and group B (dark grey) for a test set of 34 words without previous accent learning (training set size = 0) and with previous learning on a training set of 35 words. Each error bars depicts one standard error (standard deviation divided by the square root of number of samples, here 10 samples).

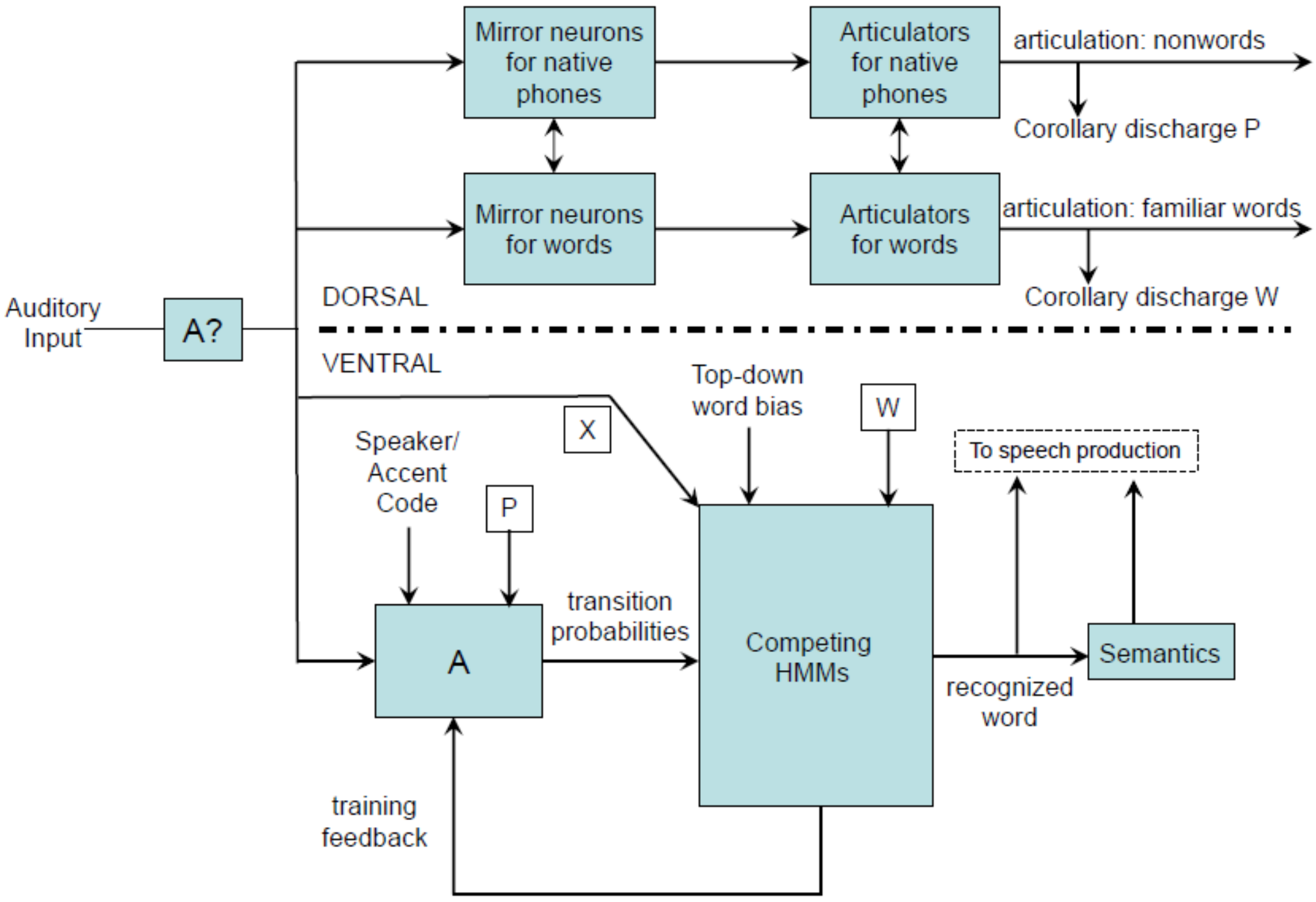

Figure 9. Embedding the accent model of Part 2 within a larger system. The dorsal pathway gives access to articulatory form, whether at the level of phoneme sequences (nonwords, or words pronounced in an unfamiliar way) or at the level of words (when they are pronounced much as the listener would). The ventral pathway gives access to semantic form which may succeed even if the word is pronounced in an accent very different from the listener's. The semantic form can access speech production (i.e., via production of the articulatory form). The "accent system" A infers likely phoneme sequences for the current auditory input if that input is in an accent it has learned, or is learning. A may receive assistance from the articulatory form of phonemes (P) registered on the dorsal path. The competition for word recognition can be based on the output of A, on fragmented cues from the auditory input (pathway X), and/or discourse cues (top-down word bias); assistance may be received from the articulatory word form (W) registered on the dorsal path. Word recognition provides training feedback to A. (The box A? indicates an alternative model, closer to the classical motor theory, in which the accent system precedes the divergence of the dorsal and ventral pathways.)

# TABLES

Table 1. Basic Notations used in the Model.

| | |
|---|---|
| N | Listener – "native" accent |
| F | Speaker – "foreign" accent |
| $\mathcal{P}$ | the set of phonemes in the native language (here, English) |
| $\mathcal{V}$ | a feature space for phones |
| $\mathcal{W}$ | a set of words of English |
| $\mathcal{X}$* | For any set $\mathcal{X}$, the set of strings of elements of $\mathcal{X}$ |
| $\Psi_S$: $\mathcal{P} \rightarrow \mathcal{V}$ | $\Psi_S$ assigns each phoneme p in the native language to the feature vector $\Psi_S(p)$ in $\mathcal{V}$ that S uses to pronounce it (the map is probabilistic in the sense that the same phoneme can be pronounced by S in different ways, depending on the context). |
| $\Phi_S$: $\mathcal{W} \rightarrow \mathcal{P}^*$ | For each word w of English in the vocabulary of S, $\Phi_S(w)$ is the string of phonemes used by S to pronounce it. |
| $\Psi_S(\Phi_S(w))$ | The sequence of feature vectors that S will produce when saying the word w slowly and clearly |
| $G_{F\rightarrow N}$: $\mathcal{V}^* \rightarrow \mathcal{W}$ | For each string θ of phones that is actually pronounced by F, $G_{F\rightarrow N}(\theta)$ is the word of English that N understands F to have intended to say by uttering this string. (The map is probabilistic.) |
| $H_{F\rightarrow N}$: $\mathcal{V} \rightarrow \mathcal{P}^*$ | For most words in $\mathcal{W}$, if $\Psi_F(\Phi_F(w))$ equals the string of feature vectors $v_1 v_2 \ldots v_k$ used by F to pronounce w, then $\Phi_N(w)$ equals (more or less) the string of phonemes $H_{F\rightarrow N}(v_1)$ $H_{F\rightarrow N}(v_2) \ldots H_{F\rightarrow N}(v_k)$. |

Table 2: Chosen transcriptions for the few words which are not in the MRC database.

| | | | |
|---|---|---|---|
| Stella: /stɛllə/ | Peas: /piz/ | Slabs: /slæbz/ | Snack: /snæk/ |
| Snake: /sneɪk/ | Toy: /tɔɪ/ | Frog: /fɹɒg/ | |

Table 3: Transformations found between the native and a French transcription of the elicitation paragraph (French8). The absence of a phoneme in the column "Native phoneme" means an insertion of the phone in the column "Foreign pronunciation", while an absence of phone in the column "Foreign pronunciation" means a deletion of the native phoneme. We count the number of occurrence of each transformation both in the learning and test sets, and compare them with the output of the model on the training set.

| Native phoneme (in $\mathcal{P}$) | Foreign pronunciation (in $\mathcal{V}$) | Learning set count (first 35 words of the elicitation paragraph) | Test set count (last 34 words of the elicitation paragraph) | Model count on learning set |
|---|---|---|---|---|
| z | z | 0 | 2 | 0 |
| z | s | 7 | 3 | 7 |
| ə | ə | 2 | 1 | 2 |
| ə | ʌ | 1 | 3 | 1 |
| ə | œ | 1 | 0 | 1 |
| ə | ə̃ | 0 | 1 | 0 |
| h | h | 1 | 1 | 1 |
| h | | 2 | 0 | 2 |
| ɜ | ɜ | 0 | 0 | 0 |
| ɜ | œ | 3 | 1 | 3 |
| ɹ | ɹ | 0 | 0 | 0 |
| ɹ | r | 1 | 0 | 0 |
| ɹ | ʀ | 3 | 4 | 0 |
| ɹ | | 0 | 0 | 4 |
| ð | ð | 0 | 0 | 0 |
| ð | d | 3 | 3 | 3 |
| ð | t | 1 | 0 | 1 |
| θ | θ | 0 | 0 | 0 |
| θ | t | 2 | 2 | 2 |
| ɒ | ɒ | 0 | 0 | 0 |
| ɒ | ʌ̃ | 1 | 0 | 1 |
| ɒ | ʌ | 2 | 0 | 2 |
| ɒ | ə | 1 | 0 | 1 |
| ɒ | ɔ | 0 | 1 | 0 |
| | ʀ | 3 | 2 | 6 |
| u | u | 2 | 1 | 2 |
| u | ũ | 1 | 0 | 1 |
| v | v | 0 | 0 | 0 |
| v | f | 3 | 0 | 3 |
| əʊ | əʊ | 0 | 0 | 0 |
| əʊ | oʊ | 1 | 2 | 1 |
| aɪ | aɪ | 0 | 0 | 0 |
| aɪ | ãɪ | 1 | 0 | 1 |
| d | d | 0 | 5 | 0 |
| d | t | 0 | 1 | 0 |
| d | | 1 | 0 | 1 |
| b | b | 4 | 2 | 4 |
| b | p | 1 | 0 | 1 |
| æ | æ | 3 | 3 | 3 |
| æ | æ̃ | 1 | 3 | 1 |
| | r | 0 | 0 | 1 |

# APPE NDIX

**Detailed word recognition for group A and group B (Figure )**

Each sample can be heard on Weinberger (2010). Two words appearing in the same brackets means that they are recognized with the same probability. Group A are native English speakers, group B are foreign speakers.

***Group A (native speakers)***

**English10**

Transcription

[p$^{h}$l̥iːs k$^{h}$ɔl stɛla æ̰̃sk ə ɾə bɹĩŋ ðiːz θĩŋz wɪθ hɚ fɹʌ̃m ðə stɔəɹ sɪks spũːnz ʌv̥ fɹɛʃ snoʊ p$^{h}$ɪːz faɪv̥ θɪk slæ̰̃ːbz ʌv̥ blu ʧiːz æ̃n meɪbiː ə snæ̃k fɚ hɚ bɹʌðɚ baːb wi ɔlso niːɾ ə smɔl p$^{h}$l̥æ̰stɪk snẽɪk ɛ̃n ə bɪg˺ t$^{h}$ʊɪ fɹaːg fə ðə k$^{h}$ɪdz̥ ʃi kæ̆n skup ðiz θĩŋz ĩnt$^{h}$ə θɹi ɹɛ̰d bæ̰̃ːgz æn wə wɪl$^{ɣ}$ goʊ miːt hɚ wɛ̃nzde æt ðə tɹẽɪn steɪʃɪn]

Word recognition on test base before learning

[ALSO - ] --- [NICE - ] --- [A - ] --- [SMALL - ] --- [PLASTIC - ] --- [SNAKE - ] --- [N - ] --- [A - ] --- [BIG - ] --- [DOING - ] --- [FROG - ] --- [THE - ] --- [THE - ] --- [KIDS - ] --- [SHE - ] --- [CAN - ] --- [SCOOP - ] --- [THESE - ] --- [THINGS - ] --- [INTO - ] --- [THREE - ] --- [RED - READ - ] --- [BAGS - ] --- [AN - ] --- [WERE - ] --- [WILL - ] --- [GO - ] --- [MEET - ] --- [HER - ] --- [WEDNESDAY - ] --- [AT - ] --- [THE - ] --- [TRAIN - ] --- [STATION - ] --- RECOGNITION RATE = 82.3529

Word recognition on test base after learning

[ALSO - ] --- [NEEDS - ] --- [A - ] --- [SMALL - ] --- [PLASTIC - ] --- [SNAKE - ] --- [N - ] --- [A - ] --- [BIG - ] --- [DOING - ] --- [FROG - ] --- [FOR - FOUR - ] --- [THE - ] --- [KIDS - ] --- [SHE - ] --- [CAN - ] --- [SCOOP - ] --- [THESE - ] --- [THINGS - ] --- [INTO - ] --- [THREE - ] --- [RED - READ - ] --- [BAGS - ] --- [AN - ] --- [WERE - ] --- [WILL - ] --- [GO - ] --- [MEET - ] --- [HER - ] --- [WEDNESDAY - ] --- [AT - ] --- [THE - ] --- [TRAIN - ] --- [STATION - ] --- RECOGNITION RATE = 85.2941

**English18**

Transcription

[pʰl̥əiz kɔl stəɛlʌ æsk hɝ tʰŭ bɹə̆ĩŋ ðiːz θɪŋz wɪθ hɝ fɹʌm ðə stoəɹ sɪks spũ̯ːnz ʌy̥ fɹɛʃ snoʊ̆ pəiːz faːv θɪk slæbz əv̌ blu̯ ʧiːz ɛ̆n mɛbi ə snæk fŏɹ hɝ bɹʌðɚ bɑːb wi also niːd ə smɑːl pʰl̥æstɪk sneɪk ɛn ə bɪg˺ tʰɔɪ fɹɔg fə ðə kʰiədz̥ ʃi kʰən sːku̯p ðiːz θəɪŋz̥ ɪntʰu θɹiː ɹed bæːgz̥ ɛn wi wɪl goʊ miːt hɝ wĩnzdi æt ðə tɹeĩn steɪʃən]

Word recognition on test base before learning

[ELSE - ] --- [NEED - ] --- [A - ] --- [SMALL - ] --- [PLASTIC - ] --- [SNAKE - ] --- [N - ] --- [A - ] --- [BIG - ] --- [TOY - ] --- [FROG - ] --- [THE - ] --- [THE - ] --- [KIDS - ] --- [SHE - ] --- [CAN - ] --- [SCOOP - ] --- [THESE - ] --- [THINGS - ] --- [INTO - ] --- [THREE - ] --- [RID - ] --- [BAGS - ] --- [N - ] --- [WE - ] --- [WILL - ] --- [GO - ] --- [MEET - ] --- [HER - ] --- [WEDNESDAY - ] --- [AT - ] --- [THE - ] --- [TRAIN - ] --- [STATION - ] ---

RECOGNITION RATE = 85.2941

Word recognition on test base after learning

[ELSE - ] --- [NEED - ] --- [A - ] --- [SMALL - ] --- [PLASTIC - ] --- [SNAKE - ] --- [N - ] --- [A - ] --- [BIG - ] --- [TOY - ] --- [FROG - ] --- [FOR - FOUR - ] --- [THE - ] --- [KIDS - ] --- [SHE - ] --- [GONE - ] --- [SCOOP - ] --- [THESE - ] --- [THINGS - ] --- [INTO - ] --- [THREE - ] --- [RID - ] --- [BAGS - ] --- [N - ] --- [WE - ] --- [WILL - ] --- [GO - ] --- [MEET - ] --- [HER - ] --- [WEDNESDAY - ] --- [AT - ] --- [THE - ] --- [TRAIN - ] --- [STATION - ] ---

RECOGNITION RATE = 85.2941

**English23**

Transcription

[pʰl̥iz kʰaʊl stɛlʌ æsk həɹ ɾŭ bɹĩŋ ðiːs̪ θĩŋz wɪθ həɹ fɹə̃m ðə̆ stoəɹ sɪks spəũːnz əv fɹɛʃ snoʊ pʰəiːz̥ faː θɪk slæːbz ə blu ʧəiːz ɛ̃n meɪbi ə sːnæk˺ fəɹ həɹ bɹʌðɚ bɑːb wi ɑlso niɾ ə smɑʊlˠ pʰl̥æstɪk sneɪk ɛ̃n ʌ bɪg tʰɔɪ fɹɑg˺ fəɹ ðə kiədz ʃi kə̃n skŭ̯p ðiːs θəĩŋz ĩntə θɹiː ɹɛ̝d baɪgz̥ æ̃n wi wəlˠ goʊ mə̆iʈ həɹ wĩnzdiː æt d̪ə tɹẹ̃ĩn steɪʃən]

Word recognition on test base before learning

[ALSO - ] --- [NICE - ] --- [A - ] --- [SMALL - ] --- [PLASTIC - ] --- [SNAKE - ] --- [N - ] --- [OR - ] --- [BIG - ] --- [TOY - ] --- [FROG - ] --- [FALL - ] --- [THE - ] --- [KIDS - ] --- [SHE - ] --- [CAN - ] --- [SCOOP - ] --- [THESE - ] --- [THINGS - ] --- [INTO - ] --- [THREE - ] --- [RED - READ - ] --- [BAGS - ] --- [AN - ] --- [WE - ] --- [WELL - ] --- [GO - ] --- [MAKE - ] --- [HELL - ] --- [WEDNESDAY - ] --- [AT - ] --- [TO - ] --- [TRAIN - ] --- [STATION - ] ---

RECOGNITION RATE = 73.5294

Word recognition on test base after learning

[NOVELS - ] --- [NICE - ] --- [A - ] --- [SMALL - ] --- [PLASTIC - ] --- [SNAKE - ] --- [AN - ] --- [A - ] --- [BIG - ] --- [TOY - ] --- [FROG - ] --- [FOR - FOUR - ] --- [THE - ] --- [KIDS - ] ---

[SHE - ] --- [GONE - ] --- [SCOOP - ] --- [THESE - ] --- [THINGS - ] --- [INTO - ] --- [THREE - ] --- [RED - READ - ] --- [BAGS - ] --- [AN - ] --- [WE - ] --- [WALL - ] --- [GO - ] --- [MEET - ] --- [HER - ] --- [WEDNESDAY - ] --- [AT - ] --- [DOOR - ] --- [TRAIN - ] --- [STATION - ] ---

RECOGNITION RATE = 79.4118

**English33**

Transcription

[pliz kɔl stɛlʌ æsk həɹ tu bɹĩŋ ðiːz θĩŋz wɪθ ɛɹ fɹʌm n̪ə stɔːɹ sɪks spũːnz ʌv fɹɛʃ ʃnoʊ piːz faɪv̥ θɪk slæːbz ə blu̞ː ʧiːz ɛ̃n meɪbi e snæk fəɹ həɹ bɹʌðəɹ bɑːb wi ɑsoʊ nid e smɔl plæsːɪk sneɪk ɛn e bɪg tɔɪ fɹɔːg fɔːɹ ðʌ kɪdz ʃi kɛ̃n sku̞ːp d̪iːz θĩŋz ĩntu̞ː θɹiː ɹɛd bæːgz ɛn wi wəl goʊ miːt ʔhəɹ wĩnzdeɪ æt ðə tɹeĩn steɪʃə̃n]

Word recognition on test base before learning

[ANSWER - ] --- [NEED - ] --- [A - ] --- [SMALL - ] --- [PLASTIC - ] --- [SNAKE - ] --- [N - ] --- [A - ] --- [BIG - ] --- [TOY - ] --- [FROG - ] --- [FALL - ] --- [THE - ] --- [KIDS - ] --- [SHE - ] --- [CAN - ] --- [SCOOP - ] --- [NICE - ] --- [THINGS - ] --- [INTO - ] --- [THREE - ] --- [RED - READ - ] --- [BAGS - ] --- [N - ] --- [WE - ] --- [WELL - ] --- [GO - ] --- [MEET - ] --- [HELL - ] --- [WEDNESDAY - ] --- [AT - ] --- [THE - ] --- [TRAIN - ] --- [STATION - ] --- RECOGNITION RATE = 79.4118

Word recognition on test base after learning

[SO - ] --- [NEED - ] --- [A - ] --- [SMALL - ] --- [PLASTIC - ] --- [SNAKE - ] --- [N - ] --- [A - ] --- [BIG - ] --- [TOY - ] --- [FROG - ] --- [FOR - FOUR - ] --- [THE - ] --- [KIDS - ] --- [SHE - ] --- [CAN - ] --- [SCOOP - ] --- [NEEDS - ] --- [THINGS - ] --- [INTO - ] --- [THREE - ] --- [RED - READ - ] --- [BAGS - ] --- [N - ] --- [WE - ] --- [WALL - ] --- [GO - ] --- [MEET - ] --- [HER - ] --- [WEDNESDAY - ] --- [AT - ] --- [THE - ] --- [TRAIN - ] --- [STATION - ] --- RECOGNITION RATE = 85.2941

**English39**

Transcription

[pliːz̥ kaʊ̆l stɛlɚ æsk hɜɹ ɾə bɹĩŋ ðiːz θĩŋz wɪθ hɜɹ fɹʌ̃m ði stɔəɹ sɪks spəũnz əv fɹɛʃ snoʊ piːz faɪv θɪk slæbz ə bluː ʧiːz ɛ̃n meɪbi ə snæk˺ fəɹ hɜɹ bɹʌðəɹ baːb wi ɔlsə niːd ə smaʊl plæ̞stɪk sneɪk ə̃n ə bɪg tɔɪ fɹɔg fə ðə kɪədz ʃi kə̃ns skuːp ðiːz θĩŋz ĩntuː t̪ɹi ɹɛd bæ̞gz ɛ̃n wi wɪl goʊ miːt hɜɹ wɛ̃nzdeɪ æt ðə tɹeĩn stẽɪʃə̃n]

Word recognition on test base before learning

[ALSO - ] --- [NEED - ] --- [A - ] --- [SMALL - ] --- [PLASTIC - ] --- [SNAKE - ] --- [AN - ] --- [A - ] --- [BIG - ] --- [TOY - ] --- [FROG - ] --- [THE - ] --- [THE - ] --- [KIDS - ] --- [SHE - ] --- [GETS - ] --- [SCOOP - ] --- [THESE - ] --- [THINGS - ] --- [INTO - ] --- [TREAT - ] --- [RED - READ - ] --- [BAGS - ] ---

[N - ] --- [WE - ] --- [WILL - ] --- [GO - ] --- [MEET - ] --- [HELL - ] --- [WEDNESDAY - ] --- [AT - ] --- [THE - ] --- [TRAIN - ] --- [STATION - ] ---

RECOGNITION RATE = 82.3529

Word recognition on test base after learning

[ALSO - ] --- [NEED - ] --- [A - ] --- [SMALL - ] --- [PLASTIC - ] --- [SNAKE - ] --- [AN - ] --- [A - ] --- [BIG - ] --- [TOY - ] --- [FROG - ] --- [FOR - FOUR - ] --- [THE - ] --- [KIDS - ] --- [SHE - ] --- [GETS - ] --- [SCOOP - ] --- [THESE - ] --- [THINGS - ] --- [INTO - ] --- [TEA - T - ] --- [RED - READ - ] --- [BAGS - ] --- [AN - ] --- [WE - ] --- [WILL - ] --- [GO - ] --- [MEET - ] --- [HER - ] --- [WEDNESDAY - ] --- [AT - ] --- [THE - ] --- [TRAIN - ] --- [STATION - ] ---

RECOGNITION RATE = 88.2353

**English40**

Transcription

[pʰl̥iːz kol stɛlʌ æ̠sk ɜ̆ tŭ bɹĩŋ ðiːz̪ t̪ĩŋz wɪð ɜː fɹə̃m n̪ə stoː sɪks spũːnz əv fɹɛʃ ʃnɛʊ pəiz fɒɪv θɪk slæ̞bz əv blṷ ʧəiz n̩ mɑɪbi ə snæk fɚ ɜ bɹʌðə bɑːb̥ wi ɔlsoʊ ne̠d ə smɔl pʰl̥æstɪk sneɪk ə̃n ə bɪg tɔɪ fɹɔ̆g fə ðə kɪd̥s ʃi kə̃n skəup ðiːz θĩŋz ĭntŭ θɹiː ɹɛd˺ bæ̠g̥s ə̃n wi wɪl go̞ʊ məit ɜ wɛ̃nzdiː ət ðə tɹeĩn stẽɪʃə̃n]

Word recognition on test base before learning

[ALSO - ] --- [DID - ] --- [A - ] --- [SMALL - ] --- [PLASTIC - ] --- [SNAKE - ] --- [AN - ] --- [A - ] --- [BIG - ] --- [TOY - ] --- [FROG - ] --- [THE - ] --- [THE - ] --- [KIDS - ] --- [SHE - ] --- [CAN - ] --- [COPE - ] --- [THESE - ] --- [THINGS - ] --- [ISN'T - ] --- [THREE - ] --- [RED - READ - ] --- [BAGS - ] --- [AN - ] --- [WE - ] --- [WILL - ] --- [GO - ] --- [BOAT - ] --- [A - ] --- [WEDNESDAY - ] --- [EAT - ] --- [THE - ] --- [TRAIN - ] --- [STATION - ] ---

RECOGNITION RATE = 73.5294

Word recognition on test base after learning

[ALSO - ] --- [THEY'D - ] --- [A - ] --- [SMALL - ] --- [PLASTIC - ] --- [SNAKE - ] --- [ON - ] --- [A - ] --- [BIG - ] --- [TOY - ] --- [FROG - ] --- [FOR - FOUR - ] --- [THE - ] --- [KIDS - ] --- [SHE - ] --- [GONE - ] --- [SCOOP - ] --- [THESE - ] --- [THINGS - ] --- [DIDN'T - ] --- [THREE - ] --- [RED - READ - ] --- [BAGS - ] --- [ON - ] --- [WE - ] --- [WILL - ] --- [GOOD - ] --- [MEET - ] --- [HER - ] --- [WEDNESDAY - ] --- [OUGHT - ] --- [THE - ] --- [TRAIN - ] --- [STATION - ] ---

RECOGNITION RATE = 79.4118

**English57**

Transcription

[plɨ̆iz koːlˠ stɛlă ask ɜ tʰə bɹɨ̆ɪŋ ðɨ̆iz θĩŋz wið a fɹɔ̃m n̪ə stɔː sɪkːs spʉ̆ũnz əf fɹɛʃ snəu pɨ̆iz fɒɪv θik slabz ə blʉ̆u tʃɨ̆iz ə̃n maɪbi ə snak fɚ ɜ bɹoðə bɔb̥ wɨ ɒlˠsəu nɨ̆id ə smolˠ pʰlastik snaɪk ə̃n ə bɪɡ̊ tʰɔɪ fɹɑɡ̊ fə ðə kits ʃi kə̆n skʉ̆up ðɨ̆iz θĩːŋz ĩntʰə θɹɨ̆i ɹɛd˺ baks ə̃nd wɨ wɪlˠ ɡəu mɨ̆it hɜ wɛ̃nzdɛɪ at ðə tɹăĩn staɪʃĩn]

Word recognition on test base before learning

[ALSO - ] --- [NEEDED - ] --- [A - ] --- [SMALL - ] --- [PLASTIC - ] --- [SNACK - ] --- [AN - ] --- [A - ] --- [BIG - ] --- [TOY - ] --- [FROG - ] --- [THE - ] --- [THE - ] --- [KIDS - ] --- [SHE - ] --- [CAN - ] --- [SCOOP - ] --- [THESIS - ] --- [THINGS - ] --- [INTO - ] --- [THREE - ] --- [RED - READ - ] --- [BAGS - ] --- [END - ] --- [WE - ] --- [WILL - ] --- [GO - ] --- [MINUTE - ] --- [HER - ] --- [WEDNESDAY - ] --- [AT - ] --- [THE - ] --- [TRAIN - ] --- [STARTED - ] ---

RECOGNITION RATE = 76.4706

Word recognition on test base after learning

[ALSO - ] --- [NEED - ] --- [A - ] --- [SMALL - ] --- [PLASTIC - ] --- [SNACK - ] --- [AN - ] --- [A - ] --- [BIG - ] --- [TOY - ] --- [FROG - ] --- [FOR - FOUR - ] --- [THE - ] --- [KIDS - ] --- [SHE - ] --- [CORNER - ] --- [SCOOP - ] --- [THESE - ] --- [THINGS - ] --- [INTO - ] --- [THREE - ] --- [RED - READ - ] --- [BANKS - ] --- [AND - ] --- [WE - ] --- [WILL - ] --- [GO - ] --- [MEET - ] --- [HER - ] --- [WEDNESDAY - ] --- [AT - ] --- [THE - ] --- [TRAIN - ] --- [STATION - ] ---

RECOGNITION RATE = 88.2353

**English58**

Transcription

[pʰliːz kʰɔːlˠ stɛlə ask ɜ tə bɹĩŋ ðiz θĩŋz wɪθ hɚ fɹʌ̃m ðə stɔːɹ sɪks spu̧nz əf fɹɛʃ snoʊ pʰiːz fɑɪv θɪk slæbz əv blu tʃiːz ə̃n mɛɪbi ă snæk fə hɜ bɹʌðɜ bɔb wi ɔlˠsoʊ nid ə smɔːlˠ pʰlæstɪk snɨ̆ɪk ə̃n ə bɪɡ̊ tʰɔɪ fɹɔɡ fə ðə kɪts ʃi kə̃n sku̧p ðiz θĩŋz ĩntə θɹi ɹɛd bagz ə̃n wi wɪlˠ ɡoʊ mit hɜ wɛ̃nzdeɪ at̪˺ ðə tʃɹeĩn steɪʃʌ̆n]

Word recognition on test base before learning

[ALSO - ] --- [NEED - ] --- [A - ] --- [SMALL - ] --- [PLASTIC - ] --- [SNAKE - ] --- [AN - ] --- [A - ] --- [BIG - ] --- [TOY - ] --- [FROG - ] --- [THE - ] --- [THE - ] --- [KIDS - ] --- [SHE - ] --- [CAN - ] --- [SCOOP - ] --- [THESE - ] --- [THINGS - ] --- [INTO - ] --- [THREE - ] --- [RED - READ - ] --- [BAGS - ] --- [AN - ] --- [WE - ] --- [WILL - ] --- [GO - ] --- [MEET - ] --- [HER - ] --- [WEDNESDAY - ] --- [AT - ] --- [THE - ] --- [TRAIN - ] --- [STATION - ] ---

RECOGNITION RATE = 91.1765

Word recognition on test base after learning

[ALSO - ] --- [NEED - ] --- [A - ] --- [SMALL - ] --- [PLASTIC - ] --- [SNAKE - ] --- [AN - ] --- [A - ] --- [BIG - ] --- [TOY - ] --- [FROG - ] --- [FOR - FOUR - ] -

-- [THE - ] --- [KIDS - ] --- [SHE - ] --- [CAN - ] --- [SCOOP - ] --- [THESE - ] --- [THINGS - ] --- [INTO - ] --- [THREE - ] --- [RED - READ - ] --- [BAGS - ] --- [AN - ] --- [WE - ] --- [WILL - ] --- [GO - ] --- [MEET - ] --- [HER - ] --- [WEDNESDAY - ] --- [AT - ] --- [THE - ] --- [TRAIN - ] --- [STATION - ] ---

RECOGNITION RATE = 94.1176

**<u>English85</u>**

<u>Transcription</u>

[pʰliz kʰɑlˠ stɛlɚ ɐsk ɚ ɾə bɹĩŋ n̪ɨiz θĩŋz wɪf hɚ fɹʌ̃m n̪ə stɔɚ səks spʉ̃nz əf fɹeʃ ʃnõ pʰɨiz faɪ θɪk slɐbz ə blʉ tʃiz ẽ mɛbi ə snæx fɚ hɚ bɹʌðɚ bɒb wĭ ɑlˠso niɾ ə smɑlˠ plɐstɪk snʲẽk ẽn ə bɛg˺ tʰɔ̝ɪ fɹɑg˺ fɚ̃ ðə kʰɪɛdz ʃi kə̃n skʉp ðiz s̪ĩŋz ĩntʉ̆ θrɨi ɹɛăd˺ bɐgz ə̃ wĭ wɪlˠ gõ mit hɚ wʲɛ̃nzde ɐt̪˺ ðə tʰɹẽn steːʃn̩]

<u>Word recognition on test base before learning</u>

[ALSO - ] --- [NICE - ] --- [A - ] --- [SMALL - ] --- [PLASTIC - ] --- [SNAKE - ] --- [AN - ] --- [A - ] --- [BEG - ] --- [TOY - ] --- [FROG - ] --- [THE - ] --- [THE - ] --- [KIDS - ] --- [SHE - ] --- [CAN - ] --- [SCOOP - ] --- [THESE - ] --- [THINGS - ] --- [INTO - ] --- [THIRTY - ] --- [RED - READ - ] --- [BAGS - ] --- [A - ] --- [WE - ] --- [WILL - ] --- [COR - ] --- [MEET - ] --- [HER - ] --- [WEDNESDAY - ] --- [AT - ] --- [THE - ] --- [TRAIN - ] --- [STATION - ] ---

RECOGNITION RATE = 79.4118

<u>Word recognition on test base after learning</u>

[ALSO - ] --- [NEEDN'T - ] --- [A - ] --- [SMALL - ] --- [PLASTIC - ] --- [SNAKE - ] --- [AN - ] --- [A - ] --- [BEG - ] --- [TOY - ] --- [FROG - ] --- [FOR - FOUR - ] --- [THE - ] --- [KIDS - ] --- [SHE - ] --- [CAN - ] --- [SCOOP - ] --- [THESE - ] --- [THINGS - ] --- [INTO - ] --- [THIRTEEN - ] --- [RED - READ - ] --- [BAGS - ] --- [A - ] --- [WE - ] --- [WILL - ] --- [GO - ] --- [MEET - ] --- [HER - ] --- [WEDNESDAY - ] --- [AT - ] --- [THE - ] --- [TRAIN - ] --- [STATION - ] ---

RECOGNITION RATE = 85.2941

**<u>English134</u>**

<u>Transcription</u>

[pʰliːs kʰɒlˠ stɛ̝lə ask ɚ tə bɹĩŋ n̪iz̥ θĩŋz wɪθ hɝ fɹʌ̃m n̪ə stɝ sɪks spʉ̝̃ːnz əv fɹɛ̝ʃ s̠nəʊ pʰiːz faɪv θɪk slæ̝b̥s əv̥ blu̝ː tʰʃiːz ãn meɪbi ə snaːk fɚ hɝ bɹʌðɚ bɒb̥ wi ɑlˠso niːd ə smɑ̣lˠ pʰlastɪk sneːɪk ãnd ə bɪg tʰɔɪ fɹɒg fə ðə kʰɪ̝dz̥ ʃi kə̃n sku̝p ðiz θĩŋz ĩntə θɹi ɹɛ̝d baːgz ɛ̃n wi wɪlˠ gəʊ mit hɚ wɛ̃də̆nzdeɪ ə̆t də tʰɹeĩn steɪʃə̃n]

<u>Word recognition on test base before learning</u>

[ALSO - ] --- [NEED - ] --- [A - ] --- [SMALL - ] --- [PLASTIC - ] --- [SNAKE - ] --- [AND - ] --- [A - ] --- [BIG - ] --- [TOY - ] --- [FROG - ] --- [THE - ] ---

[THE - ] --- [KIDS - ] --- [SHE - ] --- [CAN - ] --- [SCOOP - ] --- [THESE - ] --- [THINGS - ] --- [INTO - ] --- [THREE - ] --- [RED - READ - ] --- [BAGS - ] --- [N - ] --- [WE - ] --- [WILL - ] --- [GO - ] --- [MEET - ] --- [HER - ] --- [WEDNESDAY - ] --- [EAT - ] --- [TO - ] --- [TRAIN - ] --- [STATION - ] ---

RECOGNITION RATE = 88.2353

<u>Word recognition on test base after learning</u>

[ALSO - ] --- [NEED - ] --- [A - ] --- [SMALL - ] --- [PLASTIC - ] --- [SNAKE - ] --- [AND - ] --- [A - ] --- [BIG - ] --- [TOY - ] --- [FROG - ] --- [FOR - FOUR - ] --- [THE - ] --- [KIDS - ] --- [SHE - ] --- [CAN - ] --- [SCOOP - ] --- [THESE - ] --- [THINGS - ] --- [INTO - ] --- [THREE - ] --- [RED - READ - ] --- [BAGS - ] --- [N - ] --- [WE - ] --- [WILL - ] --- [GO - ] --- [MEET - ] --- [HER - ] --- [WEDNESDAY - ] --- [OUGHT - ] --- [DOOR - ] --- [TRAIN - ] --- [STATION - ] ---

RECOGNITION RATE = 91.1765

***Group B (foreign speakers)***

**<u>Arabic16</u>**

<u>Transcription</u>

[pliz kɑlˠ stɛlə æsx hɚ tu bɹĩŋg d̪ɪs t̪ĩŋz vɪt heɹ fɹʌ̃m də stɔ sɪks spũnz ə fɹɛs znoʊ p̬ɪz faɪf s̪ɪk slæb̥s əv blu tʃʰiz æ̃ meɪbi ə s̬næk fə hɚ bɹʌðə bɒp wi ɑlˠsoʊ nid˺ ə s̬mal plæsɪk˺ sneɪk ɛ̃nd ɛ bɪk tɔɪ fɹɑk fɔɹ də kɪz̥ ʃi kæ̃ skup˺ d̪is t̪ĩŋks ĩntu tɹi ɹɛd bæks ɛ wi wil go mit hɚ wɛ̃zdeɪ æt˺ də tʰɹeĩn steɪʃən]

<u>Word recognition on test base before learning</u>

[ALSO - ] --- [NEED - ] --- [A - ] --- [SPELL - ] --- [PLASTIC - ] --- [SNAKE - ] --- [END - ] --- [AIR - ] --- [PICK - ] --- [TOY - ] --- [FROG - ] --- [FALL - ] --- [TO - ] --- [CASE - ] --- [SHE - ] --- [CARE - ] --- [SCOOP - ] --- [NICE - ] --- [THINKS - ] --- [INTO - ] --- [TREAT - ] --- [RED - READ - ] --- [BAGS - ] --- [AIR - ] --- [WE - ] --- [WE'LL - ] --- [COR - ] --- [MEET - ] --- [HER - ] --- [WEDNESDAY - ] --- [AT - ] --- [TO - ] --- [TRAIN - ] --- [STATION - ] ---

RECOGNITION RATE = 55.8824

<u>Word recognition on test base after learning</u>

[ALSO - ] --- [NEED - ] --- [A - ] --- [SMALL - ] --- [PLASTIC - ] --- [SNAKE - ] --- [END - ] --- [N - ] --- [BIG - ] --- [TOY - ] --- [FROG - ] --- [FALL - ] --- [THE - ] --- [KIDS - ] --- [SHE - ] --- [CAN - ] --- [SCOOP - ] --- [THESE - ] --- [THINKS - ] --- [INTO - ] --- [THREE - ] --- [RED - READ - ] --- [BAGS - ] --- [N - ] --- [WE - ] --- [WE'LL - ] --- [GOOD - ] --- [MEET - ] --- [HER - ] --- [WEDNESDAY - ] --- [AT - ] --- [THE - ] --- [TRAIN - ] --- [STATION - ] ---

RECOGNITION RATE = 79.4118

**Arabic17**

Transcription

[pliz kɒlˠ stʰɛlə æsk hʌ tŭ bɹĩŋ ðɪ̠s θĩŋs wɪθ hʌ fɹʌ̃n̺ d̺ə stɔɹ sɪks spũns əf fɹɛs snoʊ p̬iz fay̥ θɪk slæb̥s əv blu tʃʰiz ɛ̃n meɪbi ə snæk fə hʌ bɹʌðə bob wɪ ɑlˠsoʊ nid̥ ə smʌlˠ plæsɪk sneɪk ɛ̃n ə bɪk tʰɔɪ fɹɒg˺ fʌ ðə kʰɪdz̥ ʃi kɛ̃n skuɸ ðɪs s̺ĩŋs ĩntu θɹi ɹɛd bæg̊s ɛ̃n wi wɪl goʊ mit hʌɹ wɛ̃ntsdeɪ æ d̺ə tɹeĩn steɪʃən]

Word recognition on test base before learning

[ALSO - ] --- [NEED - ] --- [A - ] --- [SMALL - ] --- [PLASTIC - ] --- [SNAKE - ] --- [N - ] --- [A - ] --- [PICK - ] --- [TOY - ] --- [FROG - ] --- [FOR - FOUR - ] --- [THE - ] --- [KIDS - ] --- [SHE - ] --- [CAN - ] --- [SCOOP - ] --- [THIS - ] --- [THINGS - ] --- [INTO - ] --- [THREE - ] --- [RED - READ - ] --- [BAGS - ] --- [N - ] --- [WE - ] --- [WILL - ] --- [GO - ] --- [MEET - ] --- [HELL - ] --- [WEDNESDAY - ] --- [I - EYE - ] --- [TO - ] --- [TRAIN - ] --- [STATION - ] ---

RECOGNITION RATE = 79.4118

Word recognition on test base after learning

[ALSO - ] --- [NEED - ] --- [A - ] --- [SMALL - ] --- [PLASTIC - ] --- [SNAKE - ] --- [AN - ] --- [A - ] --- [BIG - ] --- [TOY - ] --- [FROG - ] --- [FOR - FOUR - ] --- [THE - ] --- [KIDS - ] --- [SHE - ] --- [CAN - ] --- [SCOOP - ] --- [THIS - ] --- [THINGS - ] --- [INTO - ] --- [THREE - ] --- [RED - READ - ] --- [BAGS - ] --- [AN - ] --- [WE - ] --- [WILL - ] --- [GO - ] --- [MEET - ] --- [HER - ] --- [WEDNESDAY - ] --- [ADD - ] --- [THE - ] --- [TRAIN - ] --- [STATION - ] ---

RECOGNITION RATE = 88.2353

**French23**

Transcription

[plis kɔl stɛlə æsk œ t̺u bʁĩŋ ðis θĩnks wɪθ Ï fʁʌ̃m d̺ə stɔʁ̆ sɪks spuːnz əf fʁɛʃ sno pis faɪf sθɪk slæps əv blu tʃiz æ̃n meɪbi Ï snækʰ fɔʁ hɜ bʁʌd̺əʁ bÏbə wi ɔlso nid̥ ə smɔl plæstɪk sneɪk æ̃nd ə bɪk tɔɪ fʀɔg fɔʀ d̺i kɪts ʃi kæn skub d̺is θĩnks ĩntu θriː ʁɛb bægz̥ æ̃n wi wɪl go mit hɜʁ wɛ̃nzdeɪ æt d̺ə tʁẽɪn steɪʃə̃n]

Word recognition on test base before learning

[ALSO - ] --- [NEED - ] --- [A - ] --- [SMALL - ] --- [PLASTIC - ] --- [SNAKE - ] --- [AND - ] --- [A - ] --- [PICK - ] --- [TOY - ] --- [WALK - ] --- [WALK - ] --- [D - ] --- [KIDS - ] --- [SHE - ] --- [CAN - ] --- [SCOOP - ] --- [NICE - ] --- [SIX - ] --- [INTO - ] --- [THIRTY - ] --- [M - ] --- [BAGS - ] --- [AN - ] --- [WE - ] --- [WILL - ] --- [COR - ] --- [MEET - ] --- [HER - ] --- [WEDNESDAY - ] --- [AT - ] --- [TO - ] --- [TIN - ] --- [STATION - ] ---

RECOGNITION RATE = 64.7059

Word recognition on test base after learning

[ALSO - ] --- [NEED - ] --- [A - ] --- [SMALL - ] --- [PLASTIC - ] --- [SNAKE - ] --- [AND - ] --- [A - ] --- [BRING - ] --- [TOY - ] --- [FROG - ] --- [FOR - FOUR - ] --- [THE - ] --- [KIDS - ] --- [SHE - ] --- [CAN - ] --- [SCOOP - ] --- [THESE - ] --- [THINGS - ] --- [INTO - ] --- [THREE - ] --- [M - ] --- [BAGS - ] --- [AN - ] --- [WE - ] --- [WILL - ] --- [GO - ] --- [MEET - ] --- [HER - ] --- [WEDNESDAY - ] --- [AT - ] --- [THE - ] --- [TRAIN - ] --- [STATION - ] --- RECOGNITION RATE = 91.1765

**French8**

Transcription

[pl̥iːs kɔl stɛlʌ æsk œ tu brɪŋ d̪iːs t̪iŋs wɪt̪ œ fʀ̥ʌ̃mː də stɔʀ sɪks spũns ʌf fʀɛʃ snoʊ piːs fãɪf t̪ɪk slæps ʌf blu ʧiːs æ̃n meɪbi ə snæk foʀ hœ bʀʌdœʀ bəp wi olsoʊ nit ʌ smɔl plæstɪk sneɪk æ̃nd ʌ bɪk tɔɪ fʀɔg foʀ d̪ʌ kɪd̥s ʃi kæ̃n skuːb diz t̪ĩŋs ɪnt̬u tʀi ʀɛd bæg̥s æ̃nd wi wɪl goʊ miːt hœʀ wɛ̃nzde æt d̪ə tʀeĩn steɪʃə̃n]

Word recognition on test base before learning

[ALSO - ] --- [NEED - ] --- [OR - ] --- [SMALL - ] --- [PLASTIC - ] --- [SNAKE - ] --- [AND - ] --- [OR - ] --- [PICK - ] --- [TOY - ] --- [WALK - ] --- [VAGUE - ] --- [DOOR - ] --- [KIDS - ] --- [SHE - ] --- [CAN - ] --- [SCOOP - ] --- [NICE - ] --- [TAKES - ] --- [INTO - ] --- [TEA - T - ] --- [GET - ] --- [BAGS - ] --- [AND - ] --- [WE - ] --- [WILL - ] --- [GO - ] --- [MEET - ] --- [HANG - ] --- [WEDNESDAY - ] --- [AT - ] --- [TO - ] --- [TAKEN - ] --- [STATION - ] --- RECOGNITION RATE = 61.7647

Word recognition on test base after learning

[ALSO - ] --- [NEED - ] --- [A - ] --- [SMALL - ] --- [PLASTIC - ] --- [SNAKE - ] --- [AND - ] --- [A - ] --- [BIG - ] --- [TOY - ] --- [FROG - ] --- [FOR - FOUR - ] --- [THE - ] --- [KIDS - ] --- [SHE - ] --- [CAN - ] --- [SCOOP - ] --- [THESE - ] --- [THINGS - ] --- [INTO - ] --- [TEA - T - ] --- [RED - READ - ] --- [BAGS - ] --- [AND - ] --- [WE - ] --- [WILL - ] --- [GO - ] --- [MEET - ] --- [HER - ] --- [WEDNESDAY - ] --- [AT - ] --- [THE - ] --- [TRAIN - ] --- [STATION - ] --- RECOGNITION RATE = 97.0588

**Korean3**

Transcription

[plis kol stɛla æsk ɚ tu bɹĩŋ ðɪs s̪ĩŋ wɪt hɝ fɹʌ̃m d̪ə stˇɔɹ sɪks spũn ʌf fɹɛʃ snŏ piːz̥ faɪv θɪk slæb ʌf blʊ ʧiːz̥ æ̃n mebi ə snæ̃ foɹ hɝ˞ˡ bɹʌðɛɹ bab wi ɔlso nid ə smal plæsɪk sneɪk æ̆n ĕ big˺ tɔ̆ɪ fɹɔ foɹ d̪ə kits ʃi kɛ̆n skup d̪is s̪ĩŋz̥ ĩntʰɵ əθɹi ɹɛd bæːgz̥ æ̃n wi wɪl goʊ mĩd hɝ wɛ̃nəsde æt d̪ə tɹẽns steʃɪn]

Word recognition on test base before learning

[ALSO - ] --- [NEED - ] --- [A - ] --- [SPELL - ] --- [PLASTIC - ] --- [SNAKE - ] --- [AN - ] --- [A - ] --- [BIG - ] --- [TOY - ] --- [FROM - FROG - ] --- [FULL - ] --- [TO - ] --- [KIDS - ] --- [SHE - ] --- [CAN - ] --- [SCOOP - ] --- [NICE - ] --- [THINGS - ] --- [INTO - ] --- [EVERY - ] --- [RED - READ - ] --- [BAGS - ] --- [AN - ] --- [WE - ] --- [WILL - ] --- [GO - ] --- [MEET - ] --- [HER - ] --- [WEDNESDAY - ] --- [AT - ] --- [TO - ] --- [TRAINS - ] --- [STATION - ] ---

RECOGNITION RATE = 73.5294

<u>Word recognition on test base after learning</u>

[ALSO - ] --- [NEED - ] --- [A - ] --- [SMALL - ] --- [PLASTIC - ] --- [SNAKE - ] --- [AN - ] --- [A - ] --- [BIG - ] --- [TOY - ] --- [FOR - FOUR - ] --- [FALL - ] --- [THE - ] --- [KIDS - ] --- [SHE - ] --- [CAN - ] --- [SCOOP - ] --- [THESE - ] --- [THINGS - ] --- [INTO - ] --- [THREE - ] --- [RED - READ - ] --- [BAGS - ] --- [AN - ] --- [WE - ] --- [WILL - ] --- [GOOD - ] --- [MEET - ] --- [HER - ] --- [WEDNESDAY - ] --- [AT - ] --- [THE - ] --- [TRAINS - ] --- [STATION - ] ---

RECOGNITION RATE = 82.3529

**<u>Portuguese22</u>**

<u>Transcription</u>

[pliːz̪ kɔːl stɛːla ask ɝ t̪u bɹĩŋ dɪs θĩŋs wɪθə hɜɹ fɹɔ̃m də stɔːɹ sɪks spũːnz ɔf fɹɛʃ snoʊ piːs faɪf t̪ɪk s̪laːb̥z̥ ɔf blu ʃiːz ɛ̃n meɪbi ə s̬næk fɔ ɝ bɹʌːðə bɔːb wɪ ɔlso n$^{j}$iːd ə smɔl plastik$^{h}$ snak$^{h}$ ɛ̃n ə bɪg tɔɪ fɹɔːg fɔɹ də k$^{j}$iːts ʃi kɛ̃ːn skuːp dɪːz̥ s̪ĩŋz ĩntuː θɹɪ ɹɛd bæːgz ɛ̃nd wi wɪl goː mɪt hɜɹ wẽnzdeɪ æt də tɹeĩn steiʃə̃n]

<u>Word recognition on test base before learning</u>

[ALSO - ] --- [NEED - ] --- [A - ] --- [SMALL - ] --- [PLASTIC - ] --- [SNACK - ] --- [N - ] --- [A - ] --- [BIG - ] --- [TOY - ] --- [FROG - ] --- [FALL - ] --- [TO - ] --- [KIDS - ] --- [SHE - ] --- [CAN - ] --- [SCOOP - ] --- [DAYS - ] --- [THINGS - ] --- [INTO - ] --- [THREE - ] --- [RED - READ - ] --- [BAGS - ] --- [END - ] --- [WE - ] --- [WILL - ] --- [COR - ] --- [MEET - ] --- [HELL - ] --- [WEDNESDAY - ] --- [AT - ] --- [TO - ] --- [TRAIN - ] --- [STATION - ] ---

RECOGNITION RATE = 73.5294

<u>Word recognition on test base after learning</u>

[ALSO - ] --- [NEED - ] --- [A - ] --- [SMALL - ] --- [PLASTIC - ] --- [SNACK - ] --- [AN - ] --- [A - ] --- [BIG - ] --- [TOY - ] --- [FROG - ] --- [FALL - ] --- [THE - ] --- [KIDS - ] --- [SHE - ] --- [CAN - ] --- [SCOOP - ] --- [THESE - ] --- [THINGS - ] --- [INTO - ] --- [THREE - ] --- [RED - READ - ] --- [BAGS - ] --- [AND - ] --- [WE - ] --- [WILL - ] --- [GOOD - ] --- [MEET - ] --- [HER - ] --- [WEDNESDAY - ] --- [AT - ] --- [THE - ] --- [TRAIN - ] --- [STATION - ] ---

RECOGNITION RATE = 88.2353

**Spanish36**
Transcription
[pliz kʌl stɛla ɛsk ə tʃ̆u bɹĩŋ ðĭz tʰĩŋz wɪθ hɛɹ fɹʌ̃m d̪ə stɔ̞ɹ sɪks spũnz əf fɹɛʃ̟ snoʊ pʰiz faɪv θɪk slæbz ə blu tʃĭz ɛ̃n meɪbi ə snæk fə ɚ bɹʌðə bɑːb wĭ ʌlso nid ə smɑl plæstɪ s̟ːɛ̃ɪk ɛ̃n ə big˺ tʰɔɪ fɹag fʌɹ ðə kʰɪdz ʃĭ kɛ̃n skup̬ ðiz tʰĩŋz ĩntʃ̆u θri ɹɛd bægz ɛ̃n wĭ ɪl goʊ mid ɛɹ wɛ̃nzdeɪ æt̪˺ ðə tɹeĩn steɪʃən]
Word recognition on test base before learning
[ALSO - ] --- [NEED - ] --- [A - ] --- [SMALL - ] --- [PLASTIC - ] --- [SEX - ] --- [N - ] --- [A - ] --- [BIG - ] --- [TOY - ] --- [FROG - ] --- [FALL - ] --- [THE - ] --- [KIDS - ] --- [SHE - ] --- [CAN - ] --- [SCOOP - ] --- [THESE - ] --- [TAKES - ] --- [INTO - ] --- [THIRTY - ] --- [RED - READ - ] --- [BAGS - ] --- [N - ] --- [WE - ] --- [ILL - ] --- [GO - ] --- [MEET - MEAN - ] --- [S - ] --- [WEDNESDAY - ] --- [AT - ] --- [THE - ] --- [TRAIN - ] --- [STATION - ] ---

RECOGNITION RATE = 76.4706

Word recognition on test base after learning
[ALSO - ] --- [NEED - ] --- [A - ] --- [SMALL - ] --- [PLASTIC - ] --- [SEX - ] --- [AN - ] --- [A - ] --- [BIG - ] --- [TOY - ] --- [FROG - ] --- [FALL - ] --- [THE - ] --- [KIDS - ] --- [SHE - ] --- [CAN - ] --- [SCOOP - ] --- [THESE - ] --- [THINGS - ] --- [INTO - ] --- [THE - ] --- [RED - READ - ] --- [BAGS - ] --- [AN - ] --- [WE - ] --- [ILL - ] --- [GO - ] --- [MEET - ] --- [HELL - ] --- [WEDNESDAY - ] --- [AT - ] --- [THE - ] --- [TRAIN - ] --- [STATION - ] ---

RECOGNITION RATE = 79.4118

**Spanish50**
Transcription
[pʰliz̥ kɔl st̪ɛlə as hɜɹ t̪u βɹĩn̠ ði θĩns wɪθ hɜɹ fɹən d̪ə stɔːɹ sɪxs spũnz əs fɹɛʃ ʃnoʊ piːz̥ faɪf θɪk slæβ əv blu tʃis ɛ̃n meɪβi ə s̬næk fɔɹ hɜɹ bɹʌðəɹ bɑb wi ɔlso niɾ ə smɔl plæstɪ zneɪk æ̃n ə bɪk̚ tɔɪ fɹɑg fɔɹ ðə kɪts ʃi kæ̃n skuβ ði θĩns ĩntə θɹi ɹɛd bægz ɛ̃n wi wɪl goʊ mit hɜɹ wɛnɛsdeɪ at̪ d̪ə̆ tɹẽɪn steɪʃə̃n]

Word recognition on test base before learning
[ALSO - ] --- [NICE - ] --- [A - ] --- [SMALL - ] --- [PLASTIC - ] --- [SNAKE - ] --- [AN - ] --- [A - ] --- [PICK - ] --- [TOY - ] --- [FROG - ] --- [FALL - ] --- [THE - ] --- [KIDS - ] --- [SHE - ] --- [CAN - ] --- [SCOOP - ] --- [THE - ] --- [SINCE - ] --- [INTO - ] --- [THREE - ] --- [RED - READ - ] --- [BAGS - ] --- [N - ] --- [WE - ] --- [WILL - ] --- [GO - ] --- [MEET - ] --- [HELL - ] --- [WEDNESDAY - ] --- [AT - ] --- [TO - ] --- [TRAIN - ] --- [STATION - ] ---

RECOGNITION RATE = 73.5294

Word recognition on test base after learning
[ALSO - ] --- [NEEDS - ] --- [A - ] --- [SMALL - ] --- [PLASTIC - ] --- [SNAKE - ] --- [AN - ] --- [A - ] --- [BIG - ] --- [TOY - ] --- [FROG - ] --- [FOR - FOUR - ] --- [THE - ] --- [KIDS - ] --- [SHE - ] --- [CAN - ] --- [SCOOP - ] --- [THE - ] ---

[SINCE - ] --- [INTO - ] --- [THREE - ] --- [RED - READ - ] --- [BAGS - ] --- [AN - ] --- [WE - ] --- [WILL - ] --- [GO - ] --- [MEET - ] --- [HER - ] --- [WEDNESDAY - ] --- [AT - ] --- [THE - ] --- [TRAIN - ] --- [STATION - ] ---

RECOGNITION RATE = 85.2941

**Spanish54**

Transcription

[pʰliz̥ kɔːl stɛlə æs xɛɹ tu bɹĩŋ̟ d̪is̪ θĩŋks wɪθ hɛɹ fɹə̃m d̪ə stɔ sɪks spʰũnz əf fɹɛʃ sno pʰis faɪ θɪ̠k slæb̥s əv̥ blu tʃiz̥ æ̃n meɪbi ɛ̆ snæk fɝ həɹ bɹʌðɚ bɔf wi ɔlso niɾ ə smɔl plastɪk sneɪk æ̃n ə bɪ̠g̊̆ tʰɔɪ fɹɔg fəɹ ðə kɪts ʃi kæ̃n skuβ ði̪s θĩŋs ĩntə θɹi ɹɛd bæg̊s æ̃ːn wi wɪl go mit hʲɛə wɛ̃zd̪eɪ æ ðə tɹẽɪn steɪʃə̃n]

Word recognition on test base before learning

[ALSO - ] --- [NICE - ] --- [A - ] --- [SMALL - ] --- [PLASTIC - ] --- [SNAKE - ] --- [AN - ] --- [A - ] --- [BIG - ] --- [TOY - ] --- [FROG - ] --- [FALL - ] --- [THE - ] --- [KIDS - ] --- [SHE - ] --- [CAN - ] --- [SCOOP - ] --- [THESE - ] --- [THINGS - ] --- [INTO - ] --- [THREE - ] --- [RED - READ - ] --- [BAGS - ] --- [AN - ] --- [WE - ] --- [WILL - ] --- [COR - ] --- [MEET - ] --- [HAIR - ] --- [WEDNESDAY - ] --- [I - EYE - ] --- [THE - ] --- [TRAIN - ] --- [STATION - ] ---

RECOGNITION RATE = 79.4118

Word recognition on test base after learning

[ALSO - ] --- [NICE - ] --- [A - ] --- [SMALL - ] --- [PLASTIC - ] --- [SNAKE - ] --- [AN - ] --- [A - ] --- [BIG - ] --- [TOY - ] --- [FROG - ] --- [BORROW - ] --- [THE - ] --- [KIDS - ] --- [SHE - ] --- [CAN - ] --- [SCOOP - ] --- [THESE - ] --- [THINGS - ] --- [INTO - ] --- [THREE - ] --- [RED - READ - ] --- [BAGS - ] --- [AN - ] --- [WE - ] --- [WILL - ] --- [GO - ] --- [MEET - ] --- [HAIR - ] --- [WEDNESDAY - ] --- [ADD - ] --- [THE - ] --- [TRAIN - ] --- [STATION - ] ---

RECOGNITION RATE = 82.3529